\documentclass[journal]{IEEEtran}
\ifCLASSINFOpdf
   \usepackage[pdftex]{graphicx}
   \graphicspath{{../pdf/}{../jpeg/}}
   \DeclareGraphicsExtensions{.pdf,.jpeg,.png}
\else
   \usepackage[dvips]{graphicx}
   \DeclareGraphicsExtensions{.eps}
\fi
\usepackage{multirow}
\usepackage{booktabs}
\usepackage{makecell}
\usepackage{graphicx}
\usepackage{tabu}
\usepackage{lastpage}
\makeatletter
\def\hlinew#1{%
	\noalign{\ifnum0=`}\fi\hrule \@height #1 \futurelet
	\reserved@a\@xhline}
\usepackage{amsmath}
\usepackage{amsfonts}
\usepackage{fancyhdr}
\usepackage{algorithm}
\usepackage{algpseudocode}
\usepackage{url}
\begin{document}

\title{Gated-GAN: Adversarial Gated Networks for Multi-Collection Style Transfer}

\author{Xinyuan~Chen,~\IEEEmembership{}
	Chang~Xu,~\IEEEmembership{}
	Xiaokang~Yang,~\IEEEmembership{Senior~Member,~IEEE,}
	Li~Song,~\IEEEmembership{}
	and~Dacheng~Tao,~\IEEEmembership{Fellow,~IEEE}% <-this % stops a space
	\thanks{This article has been accepted for publication in a future issue of this journal, but has not been fully edited. Content may change prior to final publication.
	Citation information: DOI 10.1109/TIP.2018.2869695, IEEE Transactions on Image Processing}
	\thanks{\copyright 20XX IEEE. Personal use of this material is permitted. Permission from IEEE must be obtained for all other uses, in any current or future media, including reprinting/republishing this material for advertising or promotional purposes, creating new collective works, for resale or redistribution to servers or lists, or reuse of any copyrighted component of this work in other works.}
	\thanks{X. Chen is with the Department of Electronic Engineering, Shanghai Jiao Tong University, Shanghai 200240, China, and also with the Centre for Artificial Intelligence and the Faculty of Engineering and Information Technology, University of Technology Sydney, Ultimo, NSW 2007, Australia (e-mail: xychen91@sjtu.edu.cn).}% <-this % stops a space
	\thanks{C. Xu and D. Tao are with the UBTech Sydney Artificial Intelligence Centre and the School of Information Technologies, the Faculty of Engineering and Information Technologies, The University of Sydney, 6 Cleveland St, Darlington, NSW 2008, Australia (e-mail: c.xu@sydney.edu.au; dacheng.tao@sydney.edu.au).}% <-this % stops a space
	\thanks{X. Yang and S. Li are with the Department of Electronic Engineering, Shanghai Jiao Tong University, Shanghai 200240, China (e-mail: xkyang@sjtu.edu.cn; song\_li@sjtu.edu.cn).}
}

\markboth{}% 
{Shell \MakeLowercase{\textit{et al.}}: Bare Demo of IEEEtran.cls for IEEE Journals}

\maketitle

\begin{abstract}
 Style transfer describes the rendering of an image's semantic content as different artistic styles. Recently, generative adversarial networks (GANs) have emerged as an effective approach in style transfer by adversarially training the generator to synthesize convincing counterfeits. However, traditional GAN suffers from the mode collapse issue, resulting in unstable training and making style transfer quality difficult to guarantee. In addition, the GAN generator is only compatible with one style, so a series of GANs must be trained to provide users with choices to transfer more than one kind of style. In this paper, we focus on tackling these challenges and limitations to improve style transfer. We propose adversarial gated networks (Gated-GAN) to transfer multiple styles in a single model. The generative networks have three modules: an encoder, a gated transformer, and a decoder. Different styles can be achieved by passing input images through different branches of the gated transformer. To stabilize training, the encoder and decoder are combined as an auto-encoder to reconstruct the input images. The discriminative networks are used to distinguish whether the input image is a stylized or genuine image. An auxiliary classifier is used to recognize the style categories of transferred images, thereby helping the generative networks generate images in multiple styles. In addition, Gated-GAN makes it possible to explore a new style by investigating styles learned from artists or genres. Our extensive experiments demonstrate the stability and effectiveness of the proposed model for multi-style transfer. 

\end{abstract}


% \fi
\IEEEpeerreviewmaketitle

\section{Introduction}
\IEEEPARstart{S}{tyle} transfer refers to redrawing an image by imitating another artistic style. Specifically, given a reference style, one can make the input image look like it has been redrawn with a different stroke, perceptual representation, color scheme, or that it has been retouched using a different artistic interpretation. Manually transferring the image style by a professional artist usually takes considerable time. However, style transfer is a valuable technique with many practical applications, for example quickly creating cartoon scenes from landscapes or city photographs and providing amateur artists with guidelines for painting. Therefore, optimizing style transfer is a valuable pursuit.

Style transfer, as an extension of texture transfer, has a rich history. Texture transfer aims to render an object with the texture extracted from a different object  \cite{efros1999texture,lee2010directional,ashikhmin2003fast,hertzmann2001image}.  
In the early days, texture transfer used low-level visual features of target images, while the latest style transfer approaches are based on semantic features derived from pre-trained convolutional neural networks (CNNs).
Gatys \textit{et al.} \cite{gatys2016image} introduced the neural style transfer algorithm to separate natural image content and style to produce new images by combining the content of an arbitrary photograph with the styles of numerous well-known works of art. A number of variants emerged to improve the speed, flexibility, and quality of style transfer. 
Johnson \textit{et al.} \cite{johnson2016perceptual} and Ulyanov \textit{et al.} \cite{ulyanov2016texture}  accelerated style transfer by using feedforward networks, while Chen \textit{et al.} \cite{chen2017stylebank},  Li \textit{et al.} \cite{li2017diversified} and Odena \textit{et al.} \cite{odena2016conditional} achieved multi-style transfer by extracting each style from a single image. 
Ulyanov \textit{et al.} \cite{ulyanov2017improved} and Luan \textit{et al.} \cite{luan2017deep} enhanced the quality of style transfer by investigating instance normalization in feedforward networks  \cite{ulyanov2016texture}.

CNN-based style transfer methods can now produce high-quality imitative images. However, these methods focus on transferring the original image to the style provided by another style image (typically a painting). In contrast, collection style transfer aims to stylize a photograph by mimicking an artist's or genre's style. 
In practice, when a user takes a picture of a beautiful landscape, he might hope to re-render it on canvas such that it appears to have been painted by an artist, e.g., \textit{Monet}, or in the style of a famous animation, e.g., \textit{Your Name}. 
Given an in-depth understanding of an artist's collection of paintings, it is possible to imagine how the artist might render the scene.

With this in mind, generative adversarial networks (GANs) \cite{goodfellow2014generative} can be applied to learn the distribution of an artist's paintings. 
GANs are a framework in which two neural networks compete with each other: a generative network and a discriminative network.
The generative and discriminative networks are simultaneously optimized in a two-player game, where the discriminative networks aim to determine whether or not the input is painted by the artist, while the generative networks learn to generate images to fool the discriminative networks. However, the GAN training procedure is unstable. In particular, without paired training samples, the original GANs cannot guarantee that the output imitations contain the same semantic information as that of the input images. 
CycleGAN \cite{zhu2017unpaired}, DiscoGAN \cite{kim2017learning}, DualGAN \cite{yi2017dualgan} 
proposed cycle-consistent adversarial networks to address the unpaired image-to-image translation problem. They simultaneously trained two pairs of generative networks and discriminative networks, one to produce imitative paintings and the other to transform the imitation back to the original photograph and pursue cycle consistency. 

Considering the wide application of style transfer on mobile devices, space-saving is an important algorithm design consideration. Methods of CycleGAN \cite{zhu2017unpaired}, DiscoGAN \cite{kim2017learning}, DualGAN \cite{yi2017dualgan} could only transfer one style per network. In this work, we propose a gated transformer module to achieve multi-collection style transfer in a single network. 
Moreover, previous methods adopted cycle-consistent loss requires an additional network that converts the stylized image into the original one. With the increase of the number of transferred style, the training algorithm will become complicated if we adopt cycle-consistent loss. Also, style transfer is actually a one-sided translation problem, which does not expect style images to be transformed to content images. In our method, we adopt encoder-decoder subnetwork and an auto-encoder reconstruction loss to guarantee that the outputs have the consistent semantic information with the content images. With auto-encoder reconstruction loss, our algorithm achieves one-sided mapping, which needs less parameters and can be easily generalized for multiple styles.

\begin{figure*}
	\centering
	\includegraphics[width=0.8\linewidth]{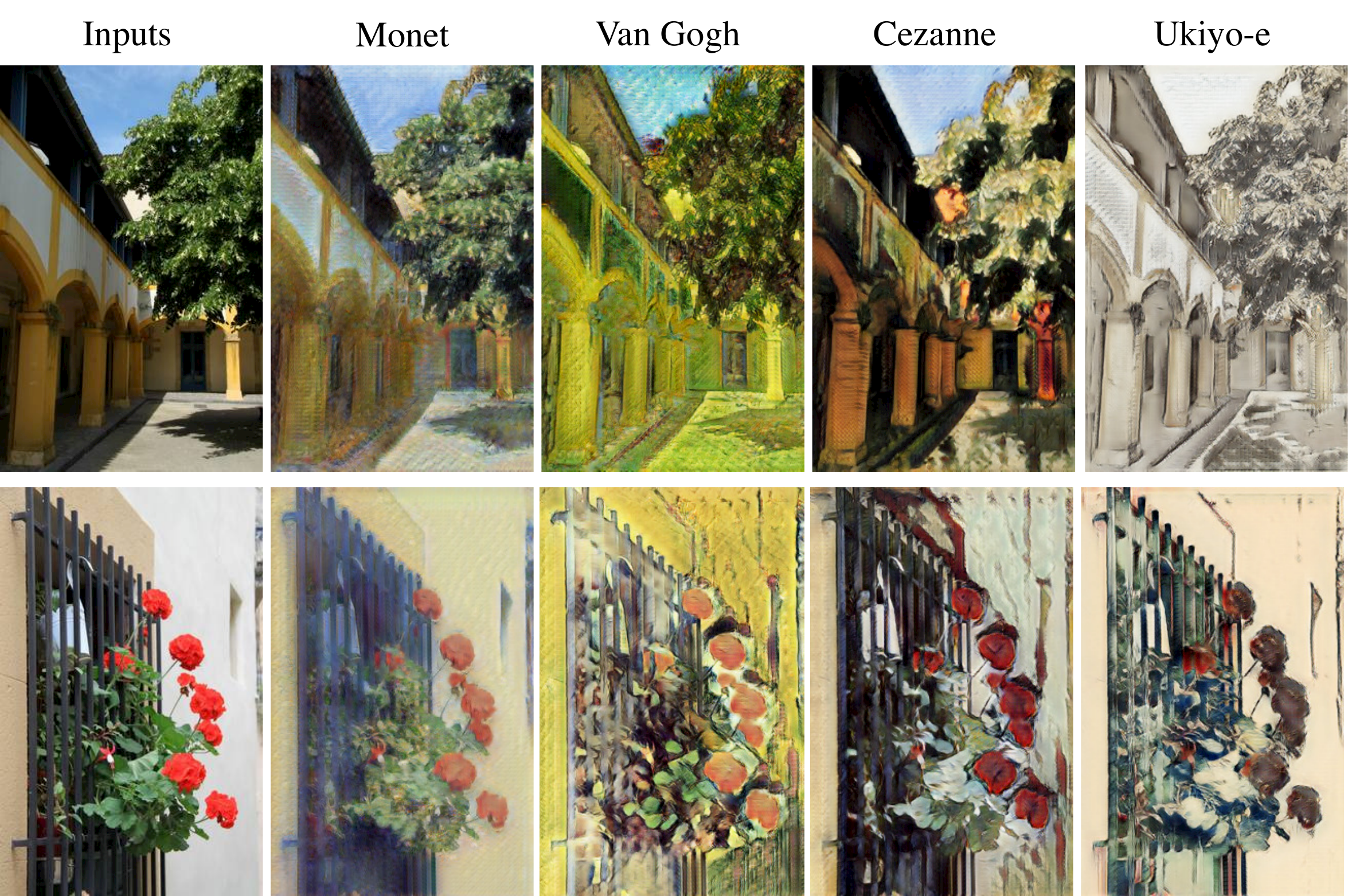}
	\caption{Gated-GAN for multi-collection style transfer. The images are produced from a single model with a shared encoder and decoder are shared. Styles are controlled by switching  different gated-transformer module. From left to right: original images, transferred images in Monet style, transferred images in Van Gogh's style, transferred images in Cezanne's style, transferred images in Ukiyoe-e's style.}
	\label{fig:multistyle}
\end{figure*}

The proposed adversarial gated networks (Gated-GAN) realize the transfer of multiple artist or genre styles in a single network (see Figure \ref{fig:multistyle}). 
Different to the conventional encoder-decoder architectures in \cite{johnson2016perceptual, radford2015unsupervised, zhu2017unpaired}, 
we additionally consider a gated-transformer network between the encoder and decoder consisting of multiple gates, each corresponding to one style. The gate controls which transformer is connected to the model so that users can switch gate to choose between different styles. If the gated transformer is skipped, the encoder and decoder are trained as an auto-encoder to preserve semantic consistency between input images and their reconstructions. At the same time, the mode collapse issue is avoided and the training procedure is stabilized. The gated transformer also facilitates generating new styles through weighted connections between the transformer branches. Our discriminative network architecture has two components: the first to distinguish synthesized images from genuine images, and the other to identify the specific styles of these images. 
Experiments demonstrate that our adversarial gated networks successfully achieve multi-collection style transfer with a quality that is better or at least comparable to existing methods.

The remainder of this paper is organized as follows. In Section 2, we summarize related work. The proposed method is detailed in Section 3. The results of experiments using the proposed method and comparisons with existing methods are reported in Section 4. We conclude in Section 5.

{\tiny }\section{Related Work}
In this section, we introduce related style transfer works. We classify style transfer methods into four categories: texture synthesis-based methods, optimization-based methods, feedforward network-based methods, and adversarial network-based methods. 

\subsection{Traditional Texture Transfer Method}
Style transfer is an extension of texture transfer, the goal of the latter being to render an object with a texture taken from a different object  \cite{efros1999texture,lee2010directional,ashikhmin2003fast,hertzmann2001image}.  
Most previous texture transfer algorithms rely on texture synthesis methods and low-level image features to preserve target image structure. Texture synthesis is the process of algorithmically constructing an unlimited number of images from a texture sample. The generated images are perceived by humans to be of the same texture but not exactly like the original images. A large range of powerful parametric and non-parametric algorithms exist to synthesize photo-realistic natural texture \cite{efros2001image,7740959,7163318}. 
Based on texture synthesis, \cite{elad2017style} and \cite{frigo2016split} used segmentation and patch matching to preserve information content. However, the texture transfer methods use only low-level target image features to inform texture transfer and take a long time to migrate a style from one image to another.

\subsection{Optimization-based Methods}
The success of deep CNNs for image classification \cite{NIPS2012_4824,szegedy2015going} prompted many scientists and engineers to visualize features from a CNN \cite{mahendran2015understanding}. 
DeepDream \cite{szegedy2015going}  was initially invented to help visualize what a deep neural network sees when given an image. Later, the algorithm became a technique to generate artworks in new psychedelic and abstract forms. Based on image representations derived from pre-trained CNNs, 
Gatys \textit{et al.} \cite{gatys2016image} introduced a neural style transfer algorithm to separate and recombine image content and style. This approach has since been improved in various follow-up papers. 
Li \textit{et al.} \cite{li2016combining} studied patch-based style transfer by combining generative Markov random field (MRF) models and the pre-trained CNNs. 
Selim \textit{et al.} \cite{selim2016painting} extended this idea to head portrait painting transfer by imposing novel spatial constraints to avoid facial deformations. Luan \textit{et al.} \cite{luan2017deep} studied photorealistic style transfer by assuming the input to output transformation was locally affine in color space. Optimization-based methods can produce high quality results but they are computationally expensive, since each optimization step requires a forward and backward pass through the pre-trained network.

\subsection{Feedforward Networks-based Methods}
Feedforward network-based methods accelerates the optimization procedure, which first iteratively optimizes a generative model and produces the styled image through a single forward pass.  Johnson \textit{et al.} \cite{johnson2016perceptual} and Ulyanov \textit{et al.} \cite{ulyanov2016texture} trained a feedforward network to quickly produce similar outputs. % to  approximate solutions.  
Based on \cite{ulyanov2016texture}, Ulyanov \textit{et al.}  \cite{ulyanov2017improved} then proposed to maximize quality and diversity by replacing the batch normalization module with instance normalization. After that, several works explored multi-style transfer in a single network. 
Dumoulin \textit{et al.} \cite{vincent2017learned} 
proposed conditional instance normalization, which specialized scaling and shifting parameters after normalization to each specific texture and allowed the style transfer network to learn multiple styles. 
Huang \textit{et al.} \cite{Xun2017arbitrary} introduced an adaptive instance normalization (AdaIN) layer that adjusted the mean and variance of the content input to match those of the style input. 
\cite{chen2017stylebank} introduced StyleBank, which was composed of multiple convolutional filter banks integrated in an auto-encoder, with each filter bank an explicit representation for style transfer.
\cite{li2017diversified} took a noise vector and a selection unit as input to generate diverse image styles. 
Although adopting different methods to achieve multi-style transfer, they all explicitly extracted style presentations from style images based on the Gram matrix \cite{gatys2016image}. 
Gram matrix based methods could do collection style transfer if they use several images as style. Though those methods are designed to transfer the style of a single image, they could also transfer the style of several images by averaging their Gram matrix statistics of pretrained deep features. On the other hand, our methods learns to output samples in the distribution of the style of a collection. 
\cite{li2017universal} achieved universal style transfer, by applying the style characteristics from a style image to content images in a style-agnostic manner. By whitening and coloring transformation, the feature covariance of content images could exhibit the same style statistical characteristics as the style images. In contrast, we are interested in the multi-collection style transfer problem. In contrast, we are interested in the multi-collection style transfer problem. A single image is difficult to comprehensively represent the style of an artist, and thus we study multi-collection style transfer to abstract the style of an artist from a collection of images.

\subsection{Adversarial Network-based Methods}
GANs \cite{goodfellow2014generative} represent a generative method using two networks, one as a discriminator and the other as a generator, to iteratively improve the model by a minimax game. 
Chuan \textit{et al.} \cite{li2016precomputed} proposed Markovian GANs for texture synthesis and style transfer, addressing the efficiency issue inherent in MRF-CNN-based style transfer \cite{li2016combining}. 
Spatial GAN (SGAN) \cite{jetchev2016texture}  successfully achieved data-driven texture synthesis based on GANs.
PSGAN \cite{bergmann2017learning} improved Spatial GAN to learn periodical textures by extending the structure of the input noise distribution.

By adopting adversarial loss, many works have generated realistic images for conditional image generation, e.g., frame prediction \cite{lotter2015unsupervised}, image super-resolution \cite{ledig2016photo} and image-to-image translation \cite{isola2016image}. However, these approaches often require paired images as input, which are expensive and hard to obtain in practice. Several studies have been conducted investigating domain transfer in the absence of paired images.
\cite{kim2017learning, yi2017dualgan, zhu2017unpaired} independently reported the similar idea of cycle-consistent loss to transform the image from the source domain to the target domain and then back to the original image. 
Taigman \textit{et al.} \cite{taigman2016unsupervised} proposed Domain Transfer Network, which employed a compound loss function, including an adversarial loss and constancy loss, to transfer a sample in one domain to an analog sample in another domain.

In contrast, some works have generated different image types from noise in a single generative network. One strategy was to supply both the generator and discriminator with class labels to produce class-conditional samples \cite{mirza2014conditional}. 
Another was to modify the discriminator to contain an auxiliary decoder network to output the class label for the training data \cite{odena2016semi,salimans2016improved} or a subset of the latent variables from which the samples were generated \cite{chen2016infogan}.
AC-GAN \cite{odena2016conditional} added auxiliary multi-class category loss to supervise the discriminator, which was used to generate multiple object types. Our work is different in that it focuses on exploring migrating different styles to content images.

\section{Proposed Algorithm}
\begin{figure*}
	\centering
	\includegraphics[width=1 \linewidth]{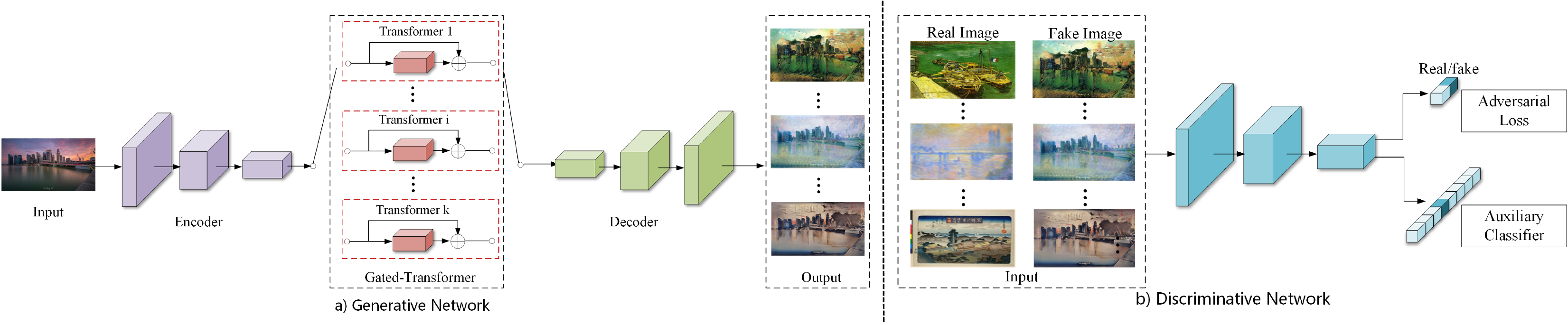}
	\caption{Architecture of the proposed adversarial gated networks: a generative network and a discriminative network. The generative network consists of three modules: an encoder, a gated transformer, and a decoder. Images are generated to different styles through branches in the gated transformer module. The discriminative network uses adversarial loss to distinguish between stylized and real images. An auxiliary classifier supervises the discriminative network to classify the style categories.}
	\label{fig:adv-dis-network}
\end{figure*}
We first consider the collection style transfer problem. We have two sets of unpaired training samples: one set of input images $ \left\{ x_i \right\} _{i=1}^N\in X$ and the target set of collections for artist or genre $ \left\{ y_i \right\} _{i=1}^M\in Y$.  We aim to train a generative network that generates images $G(x)$ in the style of a target artist or genre, and simultaneously we train a discriminative network $D$ to distinguish the transferred images $G(x)$ from the real style image $y$. The generative network implicitly learns the target style from \textit{adversarial loss}, aiming to fool the discriminator. 
The whole framework has three modules: an encoder, a gated-transformer and a decoder. The encoder consists of a series of convolutional layers that transform input image into feature space $Enc(x)$. After the encoder, a series of residual networks \cite{he2016deep} become the transformer: $T(\cdot)$. The input of residual layer in gated function $T$ is the feature maps from the last layer of encoder module $Enc(x)$. The output of the gated function is the activations $T(Enc(x))$. Then, a series of fractionally-strided convolutional networks decode the transformed feature into output images $G(x)=Dec(T(Enc(x)))$. 
To stabilize training, we introduce the \textit{auto-encoder reconstruction loss}.
We introduce the gated transformer module to integrate multiple styles within a single generated network. 
The network architecture is shown in Figure \ref{fig:adv-dis-network}, and
 the overall architecture is called the adversarial gated network (Gated-GAN).

\subsection{Adversarial Network for Style Transfer} \label{adversarial4style}
To learn a style from the target domain $Y$, we apply adversarial loss \cite{goodfellow2014generative}, which simultaneously trains $G$ and $D$ as the two-player minimax game with loss function $L(G,D)$.
The generator $G$ tries to generate an image $G(x)$ that looks similar in style to target domain $Y$, while the discriminator $D$ aims to distinguish between them. 
Specifically, we train $D$ to maximize the probability of assigning the correct label to target image $y$ and transferred image $G(x)$, meanwhile training $G$ to minimize the probability of the discriminator assigning the correct label to transferred image $G(x)$. 
The original generative adversarial value function is expressed as follows:
\begin{equation}\label{eq:adversarial loss}
\begin{split}
\min_{G} \max_{D} V(G,D) =& \mathbb{E}_{y \in Y}\left[\log D(y)\right] \\+& \mathbb{E}_{x\in X} \left[\log( 1-D(G(x)))\right] .
\end{split}
\end{equation}

We employ the least squares loss (LSGAN) as explored in \cite{mao2016least}, which provides a smooth and non-saturating gradient in the discriminator $D$. The adversarial loss $\mathcal{L}_{GAN}(G,D)$ becomes:
\begin{equation}\label{eq:lsgan}
\resizebox{.9\hsize}{!}{$
 \mathcal{L}_{GAN}(G,D) = \mathbb{E}_{y\in Y} \left[(D(y)-1)^2\right] + \mathbb{E}_{x\in X}\left[D(G(x))^2\right].
 $}
\end{equation}

\subsection{Auto-encoder Reconstruction Loss for Training Stabilization}
The original GAN framework is known to be unstable, as it must train two neural networks with competing goals. 
 \cite{zhu2017unpaired} pointed out that one reason for instability is that there exist non-unique solutions when the generator learns the mapping function. Due to unpaired training samples, the same set of input images can be mapped to any random permutation of images in the target domain. To reduce the space of possible mapping functions, we introduce the \textit{auto-encoder reconstruction loss}. In our model, the auto-encoder is obtained by directly connecting the encoder and decoder modules. That is, the network is encouraged to produce output $Dec(Enc(x))$ identical to input image $x$ after learning the representation (encoding: $Enc(x)$) for the input data. 
We define the L1 loss between the reconstructed output and input as the \textit{auto-encoder reconstruction loss}:
\begin{equation}\label{autoencoder loss}
 \mathcal{L}_R= \mathbb{E}_{x\in X}\left[||Dec(Enc(x))-x||_1\right].
\end{equation}
Mode collapse is a common problem in vanilla GAN \cite{arjovsky2017towards}, where all input images might be mapped to the same output image, and the optimization fails to make progress. In collection style transfer, if the networks trained with adversarial loss alone have sufficient capacity, content images would be mapped to an arbitrary output as long as it matches the target style. The proposed encoder-decoder subnetwork aims to reconstruct input images, so that structures of the output are expected to be consistent with the input image, which guarantees diversity of the output along with different inputs.
\subsection{Adversarial Gated Network for Multi-Collection Style Transfer}
\subsubsection{Gated Generated Network}
In multi-collection style transfer, we have a set of input images $ \left\{ x_i \right\} _{i=1}^N\in X$ and collections of paintings $Y=\left\{Y_1, Y_2,...,Y_K\right\}$, where $K$ denotes number of collections. In each collection, we have $M_c$ numbers of images  $\left\{ y_{i}\right\}_{i=1}^{M_c} \in Y_c$, where $c$ indicates the index of collection.
The proposed gated generative network aims to output images $G(x, c)$ by assigning specific style $c $. Specifically, the gated-transformer (red blocks in Figure \ref{fig:adv-dis-network}) 
transforms the input from encoded space into different styles by switching trigger to different branches: 
\begin{equation}\label{eq:G}
G\left( x,c\right)  = Dec\left( T\left(Enc(x),c \right) \right).
\end{equation}
 In each branch, we employ the residual network as the transfer module. The encoder and decoder are shared by different styles, so the network only has to save the extra transformer module parameters for each style.

\subsubsection{Auxiliary Classifier for Multiple Styles}
If we only use the adversarial loss, the model tends to confuse and mix multiples styles together. Therefore, we need a supervision to separate categories of styles. One solution is to adopt LabelGAN \cite{salimans2016improved}. \cite{salimans2016improved} generalized binary discriminator to multi-class case with its associated class label $c\in\left\{1,\cdots, K\right\}$, and the ($K+1$)-th label corresponds to the generated samples. The objective functions are defined as:
\begin{equation}\label{eq:labelgan}
\mathcal{L}_G^{lab}=\mathbb{E}_{x\in X}\left[ H\left([1,0], \left[D_r\left(G(x)\right), D_{K+1}\left(G(x)\right)\right]\right) \right],
\end{equation}
\begin{equation}
\resizebox{.91\hsize}{!}{$
\mathcal{L}_D^{lab}=\mathbb{E}_{(y,c)\in Y}\left[H(v(c),D(y)\right]+\mathbb{E}_{x \in X}\left[H(v(K+1),D(G(x))\right]
$}
\end{equation}
where  denotes the probability of the sample $x$ to have the $i$-th style. $D(x)=[D_1(x), D_2(x), \cdots, D_{K+1}(x)]$ and $v(c)=[v_1(c), …, v_{K+1}(c)]$ with $v_i(c)=0$ if $i \neq c$ and $v_i(c)=1$ if $i=c$. $H$ is the cross-entropy, defined as $H(p, q)=-\sum_i p_i \log q_i$.
In LabelGAN, the generator gets its gradients from the $K$ specific real class logits in discriminator and tends to refine each sample towards being one of the classes. However, LabelGAN actually suffers from the overlaid-gradient problem \cite{zhou2018activation}: all real class logits are encouraged at the same time. Though it tends to make each sample be one of these classes during the training, the gradient of each sample is a weighted averaging over multiple label predictors.

In our method, an auxiliary classifier (denoted as $C$) is added in the consideration of leveraging the side information directly:
\begin{eqnarray}\label{eq:L_gated}
\mathcal{L}_G^{Gated}=\lambda_{CLS}\mathbb{E}_{x\in X}\left[H(u(c), C(G(x,c))\right] + \mathcal{L}_{GAN}
\end{eqnarray}
where $u(\cdot)$ is the vectorizing operator that is similar to $v(\cdot)$ but defined with $K$ classes, and $C(G(x,c))$ is the probability distribution over $K$ real classes given by the auxiliary classifier. $L_{GAN}$ indicates the adversarial loss (in Equation \ref{eq:lsgan}) that encourages to generate realistic images. In the first term of Equation \ref{eq:L_gated}, we optimize entropy to make each sample have a high confidence of being one of the classes, so that the overlaid-gradient problem can be overcome. The loss can be written in the form of log-likelihood: 
\begin{equation}\label{ac-loss-D}
\min_C \mathcal{L}_{CLS}(C)=-\mathbb{E}_{(y,c) \in Y}\left\{\log C(Style=c|y)\right\}.
\end{equation}
The classifier $C$ is encouraged to correctly predict the log-likelihood of the correct class given real images. 
Meanwhile, the generator aims to generate images that can be correctly recognized by classifier:
\begin{equation}\label{ac-loss-G}
\min_G \mathcal{L}_{CLS}(G)=-\mathbb{E}_{x \in X}\left\{\log C(Style=c|G(x,c))\right\}.
\end{equation}
In practice, 
the classifier shares low-level convolutional layers with the discriminator, but they have exclusive fully connected layers to output the conditional distribution.

\section{Implementation}
\subsubsection{Network Configuration}
Our generative network architecture contains two stride-2 convolutions (encoder), one gated residual blocks (gated-transfer), five residual blocks, and two fractionally-convolutions with $\frac{1}{2}$ stride (decoder). Instance normalization \cite{salimans2016improved} is used after the convolutional layers. Details are provided in Table \ref{table:generative networks}.

\begin{table*}[]
	\centering
	\caption{Genrative Network of Gated-GAN}
	\label{table:generative networks}
	\begin{tabular}{lcccccc}
		\toprule
		& Operation              & Kernel size & Stride & Feature maps & Normalization          & Nonlinearity \\ \hline
		\multirow{3}{*}{Encoder} & Convolution            & 7           & 1      & 32           & Instance Normalization & ReLU         \\
		& Convolution            & 3           & 2      & 64           & Instance Normalization & ReLU         \\
		& Convolution            & 3           & 2      & 128          & Instance Normalization & ReLU         \\ \hline
		Gated-transformer        & Residual block         &             &        & 128          & Instance Normalization & ReLU         \\ \hline
		\multirow{8}{*}{Decoder} & Residual block         &             &        & 128          & Instance Normalization & ReLU         \\
		& Residual block         &             &        & 128          & Instance Normalization & ReLU         \\
		& Residual block         &             &        & 128          & Instance Normalization & ReLU         \\
		& Residual block         &             &        & 128          & Instance Normalization & ReLU         \\
		& Residual block         &             &        & 128          & Instance Normalization & ReLU         \\
		& Fractional-convolution & 3           & 1/2    & 64           & Instance Normalization & ReLU         \\
		& Fractional-convolution & 3           & 1/2    & 32           & Instance Normalization & ReLU         \\
		& Convolution            & 7           & 1      & 3            & -                      & tanh         \\ 
		\bottomrule
	\end{tabular}
\end{table*}

For the discriminators and classifiers, we adapt the Markovian Patch-GAN architecture \cite{li2016precomputed,isola2016image,zhu2017unpaired,yi2017dualgan}. Instead of operating over the full images, the discriminators and classifiers distinguish overlapping patches, sampling from the real and generated images. By doing so, the discriminators and classifiers focus on local high-frequency features like texture and style and ignore the global image structure. The patch size is set to $70 \times 70$. 
In addition, PatchGAN has fewer parameters and can be applied to any size of input.
\subsubsection{Training Strategy}
To smooth the generated image $G(x,c)$, we make use of the \textit{total variation loss} \cite{johnson2016perceptual,rudin1992nonlinear,aly2005image}, denoted by $\mathcal{L}_{TV}$: 
\begin{equation}\label{tv_loss}
\resizebox{.98\hsize}{!}{$
\mathcal{L}_{TV}=\sum_{i,j}\left[ \left(G(x)_{i,j+1}-G(x)_{i,j}\right)^2  + \left(G(x)_{i+1,j}-G(x)_{i,j}\right)^2 \right]^{\frac{1}{2}}
$}
\end{equation}
where $i\in(0,\cdots,H-1)$ and $j\in(0,\cdots,W-1)$ and $G(x)$ is the generated image whose dimension is $H \times W$.
The full objective of the generator is minimizing the loss function:
\begin{equation}\label{full-G}
\mathcal{L}(G)=\mathcal{L}_{GAN}+\lambda_{CLS}\mathcal{L}_{CLS}+\lambda_{TV}\mathcal{L}_{TV} 
\end{equation}
where $\lambda_{CLS}$ and $\lambda_{TV}$ are parameters that control relative importance of their corresponding loss functions. 
Alternatively, we train an auto-encoder by minimizing the weighted reconstruction loss in Equation \ref{autoencoder loss}: $\lambda_R\mathcal{L}_R$.
The discriminator maximizes the prediction of real images and generated images
$\mathcal{L}(D)=\mathcal{L}_{GAN}$, while the classifier in Equation \ref{ac-loss-D} maximizes the prediction of collections from different artists or genres. 
For all experiments, we set $\lambda_{CLS}=1$,
 $\lambda_R=10$, and $\lambda_{TV}=10^{-6}$. The networks are trained with a learning rate of 0.0002, using the Adam solver \cite{kingma2014adam} with batch size of 1. 
 
The input image is $128\times 128$. The training samples are first scaled to 143 $\times$ 143, and then randomly flipped and cropped to 128 $\times$ 128. We train our model with input size of 128 $\times$ 128 based on two reasons. First, randomly cropping raw input could augment the number of training set. Secondly, a relatively smaller size of image decreases the computational cost, so that speeds up training procedure. In test phase, We test images with their original resolution to receive a clearer exhibition in the paper.

To stabilize training, we update the discriminative networks using a history of transferred images rather than the ones produced by the latest generative network \cite{Shrivastava_2017_CVPR}. Specifically, we maintain an image buffer that stores 50 previously generated images. At each iteration of discriminator training, we compute the discriminator loss function by sampling images from the buffer.
The training process is shown in Algorithm \ref{alg:Framwork}. 
$\theta_{Enc}$ denotes the parameter of encoder module and $\theta_{Dec}$ denotes the parameters of decoder module.
In practice, $K_g$ and $K_d$ are set to 1.

\begin{algorithm}[htb]
	\caption{ Adversarial training of gated network $G$.}
	\label{alg:Framwork}
	\begin{algorithmic}[1]
		\Require
		The set of training sample $ \left\{ x_i \right\} _{i=1}^N\in X$,
		The set of style images with category $\left\{y_i,c_i\right\}\in Y$,
		number of discriminator network updates per step $K_d$, 
		number of generative network updates per step $K_g$.
%		Ensemble of classifiers on former batches, $E_{n-1}$;
		\Ensure		
		Gated generative newtworks: \newline 
		\centerline{$G= Dec(T(Enc(\cdot),\cdot))$.}
		\For{number of training iterations}	
			\For {$K_d$ steps}
				\State Sample minibatch of style images $(y_i,c_i)$ and training images $x_i$.
				\State Generate stylized image $G(x_i,c_i)$ in  Equation \ref{eq:G}.
				\State Update discriminator $D$ and classifier $C$ \newline
					\centerline{$\Delta_{\theta_D} \leftarrow \nabla_{\theta_{D}}\mathcal{L}_{GAN}$, $\Delta_{\theta_C} \gets {\theta_D}\nabla_{\theta_{C}}\mathcal{L}_{CLS}.$}
			\EndFor
			\For {$K_g$ steps} 
				\State Sample training images $x_i$
				\State Update generator $G$ : \newline
					$\Delta_{\theta_G}\gets \nabla_{\theta_G} 
					(\mathcal{L}_{GAN}+\lambda_{CLS}\mathcal{L}_{CLS}+\lambda_C\mathcal{L}_C+\lambda_{TV}\mathcal{L}_{TV})$.
			\EndFor
				\State Update encoder and decoder module $\theta_{Enc},\theta_{Dec}$:\newline
					\centerline{$\Delta_{\theta_{Enc},\theta_{Dec}} \gets \nabla_{\theta_{Enc},\theta_{Dec}} (\lambda_R \mathcal{L}_R)$.}
		\EndFor
	\end{algorithmic}
\end{algorithm}

\section{Experiments}
\begin{figure*}[!h]
	\centering
	\includegraphics[width=1\textwidth]{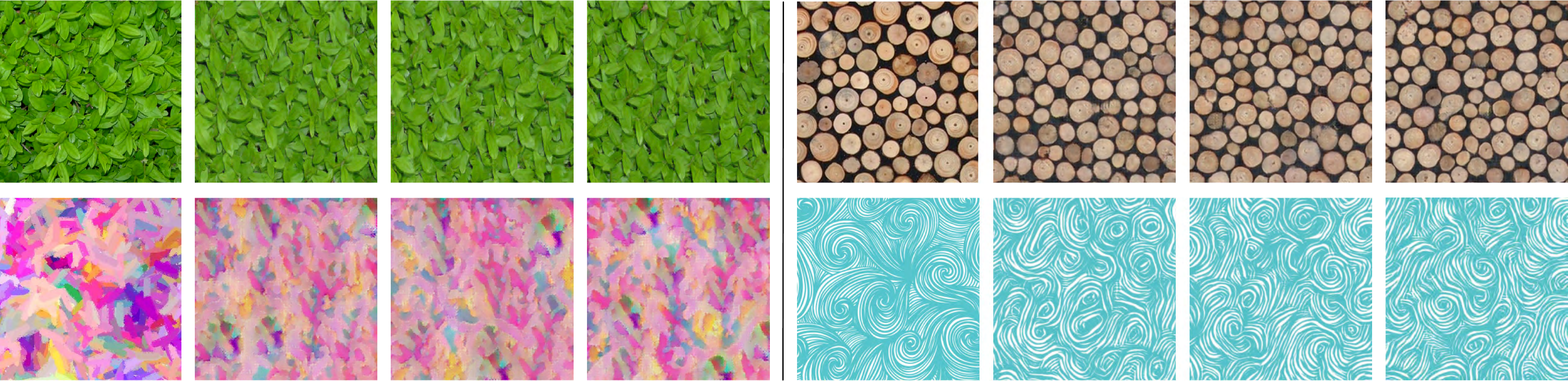}
	\caption{Four cases of texture synthesis using Gated-GAN. For each case, the first column shows examples of texture, and the other three are synthesized results given different samples of Gaussian noise as inputs.}
	\label{fig:texture}
\end{figure*}
\begin{figure*}[!htb]
	\centering
	\includegraphics[width=0.7\linewidth]{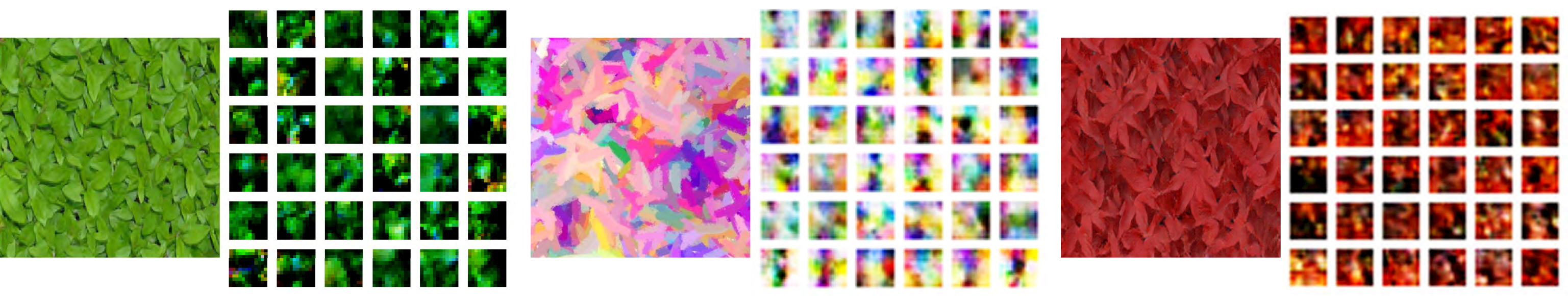}
	\caption{Visualization of learned features in the gated transformer of the generative networks. In each case, the left shows synthesized images and the right shows the corresponding features.}
	\label{fig:filters}
\end{figure*}
In this section, we evaluate the effectiveness, stability, and functionality of the proposed model. We first introduce a quantitative assessment of image quality. Then, we set up a texture synthesis experiment and visualize the filters in the gated transformer branches.  Lastly, we train the model for multiple style transfer and compare results with state-of-the-art algorithms.
\subsection{Assessment of Image Quality}
We used FID score \cite{heusel2017gans} to quantitatively evaluate the quality of results. FID score measures the distance between the generated distribution and the real distribution. To this end, the generated samples are first embedded into a feature space given by (a specific layer) of Inception Net. Then, taking the embedding layer as a continuous multi-variate Gaussian, the mean and covariance are estimated for both the generated data and the real data. The Fréchet distance between these two Gaussians is then used to quantify the quality of the samples, i.e.,
\begin{equation}\label{key}
FID(x,g)=\|\mu_x - \mu_g\|_2^2+Tr(\Sigma_x+\Sigma_g-2(\Sigma_x\Sigma_g)^{\frac{1}{2}})
\end{equation}
where $(\mu_x,\Sigma_x)$ and $(\mu_g,\Sigma_g)$ are the mean and covariance of sample embeddings from the real data distribution and generative model distribution, respectively. In our experiment, we use paintings of artists as samples of real distribution and stylized images as samples of generated distribution. That is to say, we compute the FID between generated images and authentic work of painting. 

\subsection{Texture synthesis}\label{sec:toy_exp}
To explicitly understand the gated-transformer module in the proposed Gated-GAN, we design an experiment to
explore what the gated-transformer learns.
We use our Gated-GAN to achieve synthesize texture, and visualize the gated-transformer filters. For each style, the training set is a textured image. 
The training samples are first scaled to $143 \times 143$, and then randomly flipped and cropped to $128 \times 128$.  
The generative network input is Gaussian noise. 
After adversarial training, 
the generative network outputs realistic textured images (see Figure \ref{fig:texture}).

To explore style representations learned from the gated-transformer, we visualize the transformer filters  in Figure \ref{fig:filters}.
The features are decoded by $3\times 3 \times  128$ tensors, where only one of the 128 channels is activated by Gaussian noise. 
They passed through different gated transformer filters but the same decoder. Since the output of decoder contains three channels (RGB channels), we are able to observe the colour of output decoded from the learned feature. This reveals that the transformer module learns style representations, e.g., color, stroke, etc. Another interpretation is that the transformer module learns the bases or elements of styles. Generated images can be viewed as linear combinations of these bases, with coefficients learned 
from the encoder module.

\subsection{Style Transfer}
In this subsection, we present our results for generating multiple styles of artists or genres using a single network. Then, we compare our results with state-of-the-art image style transfer and collection style transfer algorithms. The model is trained to generate images in style of Monet, Van Gogh, Cezanne, and Ukiyo-e, whose datasets are from \cite{zhu2017unpaired}. Each contains 1073, 400, 526, and 563 paintings, respectively.

\subsubsection{Multi-Collection Style Transfer}
\begin{figure}[]
	\centering
	\includegraphics[width=1\linewidth]{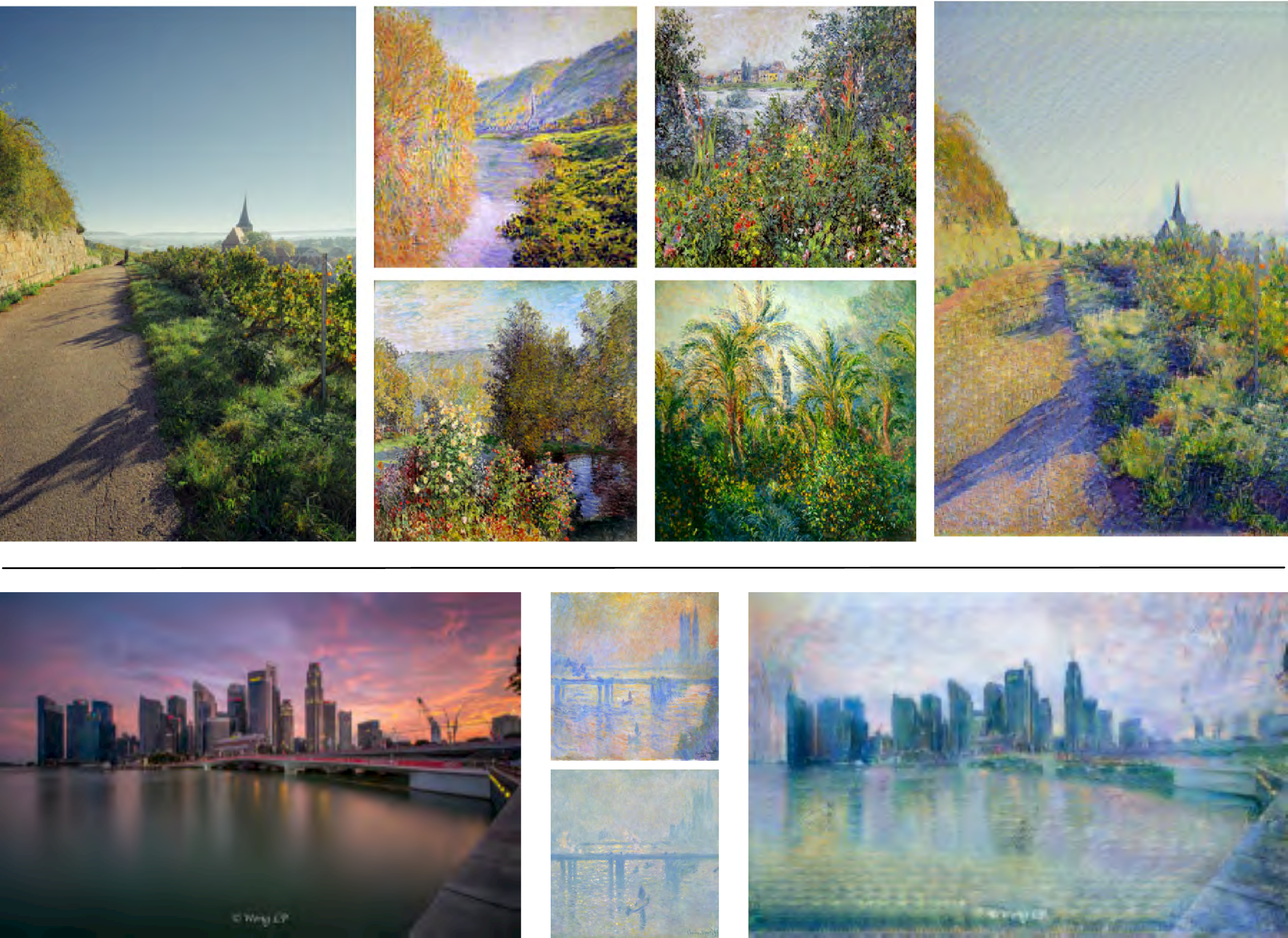}
	\caption{Collection style transfer on Photo $\rightarrow$ Monet. From left to right: input photos, Monet’s paintings picked from a similar landscape theme, and our stylized images. The photo is transferred adaptively based on different themes. }
	\label{fig:monet}
\end{figure}

\begin{figure}[!t]
	\centering
	\includegraphics[width=1\linewidth]{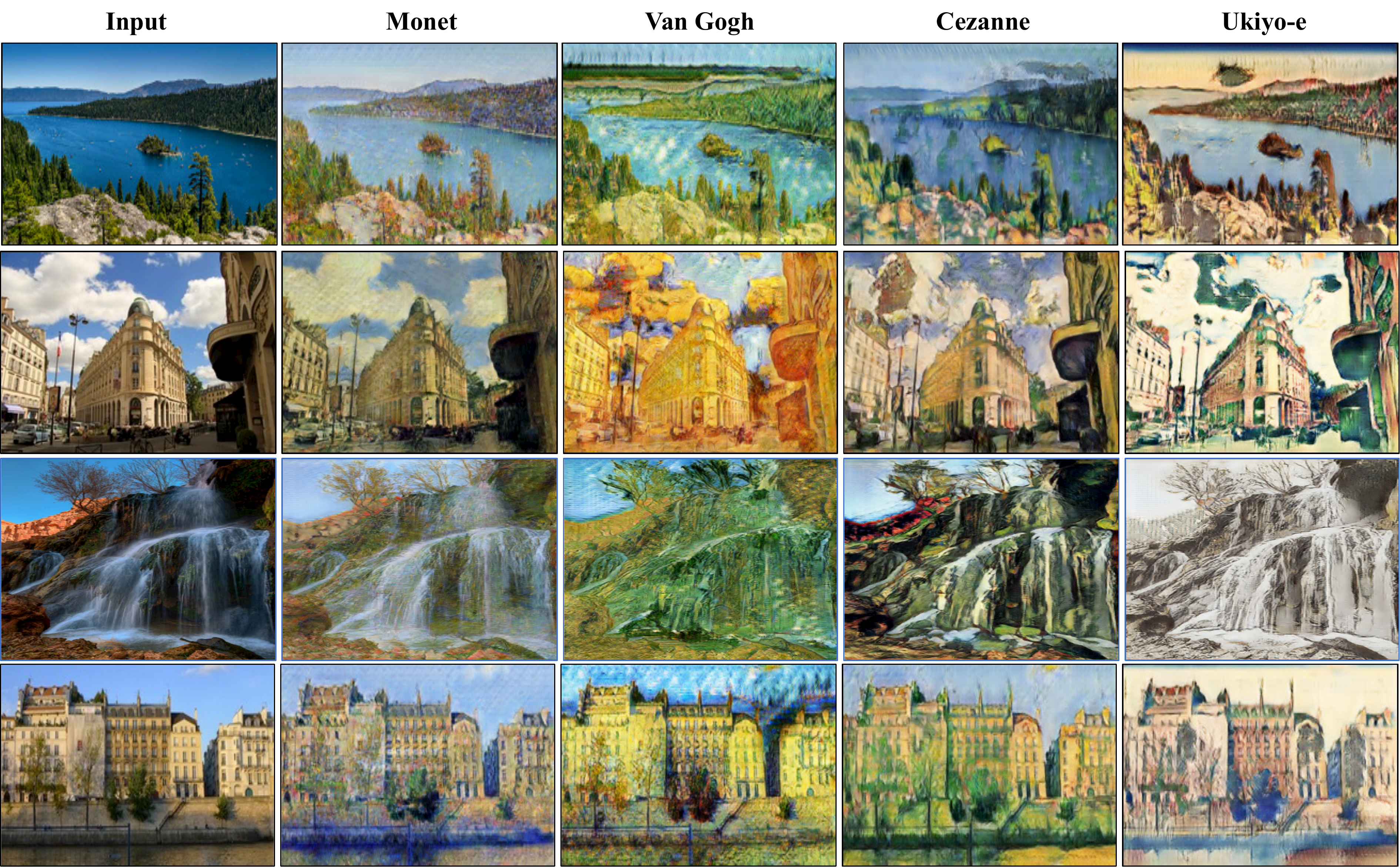}
	\caption{A four-style transfer network is trained to capture the styles of Monet, Van Gogh, Cezanne, and Ukiyo-e.} % It shows that CycleGAN and our results preserve color variation to some extent.}
	\label{fig:vscyclegan}
\end{figure}

Collection style transfer mimics the style of artists or genres with respect to their features, e.g., stroke, impasto, perspective frame usage, etc. 
Figure \ref{fig:monet} shows the results of collection style transfer using our method. Original images are presented on the left, and the generated images are on the right. For comparison, Monet’s paintings depicting similar scenes are shown in the middle. It can be seen that the styles of the generated images and their corresponding paintings are similar. Although the themes and colors of the two generated images are different, they still appear similar to Monet’s authentic pieces. Our method can clearly mimic the style of the artist for different scenes. Figure \ref{fig:vscyclegan} shows the results of applying the trained network on evaluation images for Monet’s, Van Gogh’s, Cezanne’s, and Ukiyo-e’s styles.


\subsubsection{Comparison with Image Style Transfer}
The image style transfer algorithm \cite{gatys2016image} focuses on producing images that combine the content of an arbitrary photograph and style of one or many well-known artworks. 
This is achieved by minimizing the mean-squared distance between the entries of the Gram matrix from the style image and the Gram matrix of the image to be generated. We note some recent works on multi-style transfer \cite{vincent2017learned,chen2017stylebank,li2017diversified}, but these are all based on neural style transfer \cite{gatys2016image}. Thus, we compare our results with \cite{gatys2016image}.

For each content image, we use two representative artworks as the reference style images. To generate images in the style of the entire collection, the target style representation is computed by the average Gram matrix of the target domain. To compare this with our method, we use the collections of artist’s artworks or a genre and compute the ’average style’ as the target.

\begin{figure*}
	\centering
	\includegraphics[width=0.9\linewidth]{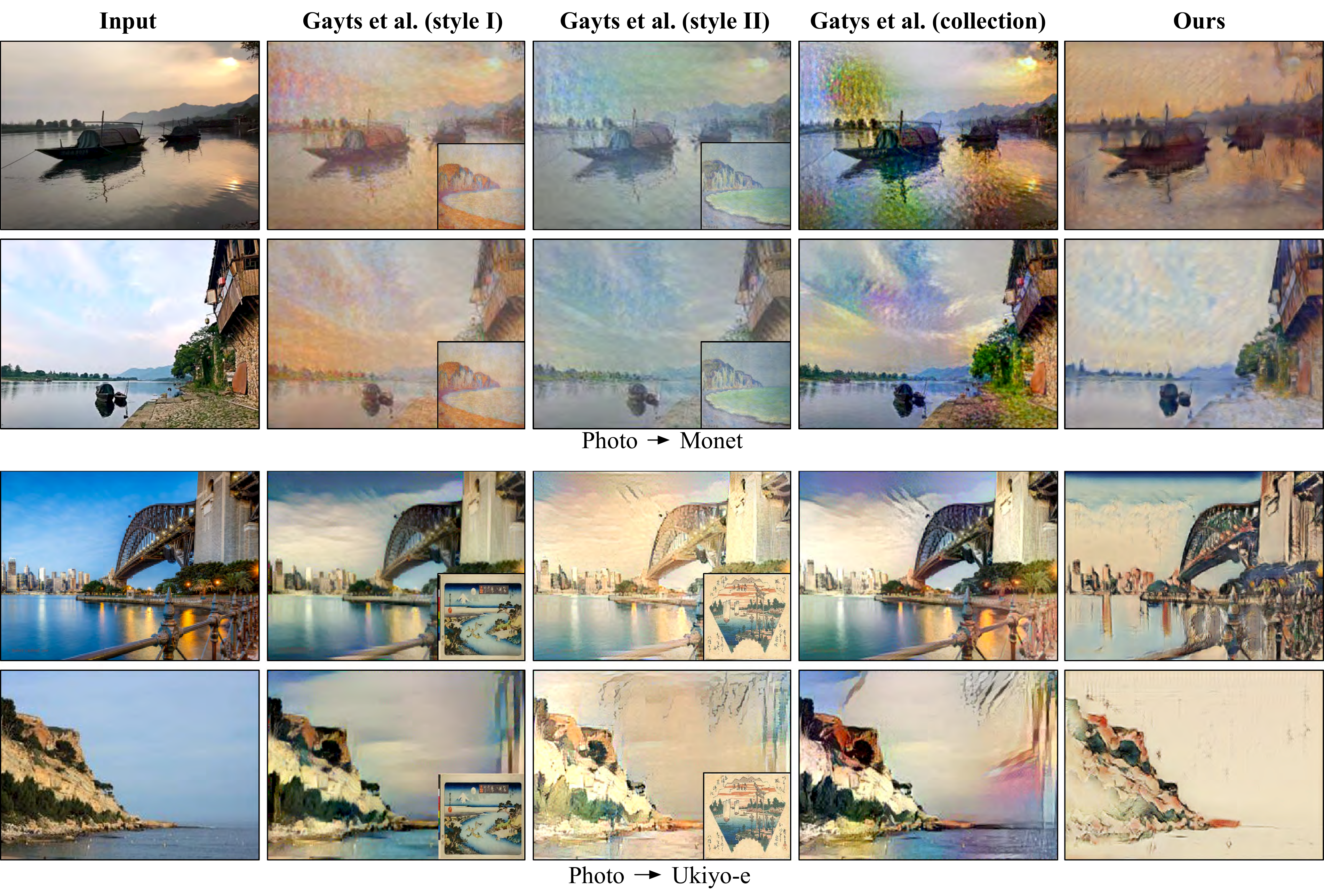}
	\vspace{-0.5cm}
	\caption{Comparison of our methods with image style transfer \cite{gatys2016image} on photo $\rightarrow$ Monet and photo $\rightarrow$ Ukiyo-e. From left to right: input photos, Gatys \textit{et al.}'s results using different target style images,  Gatys \textit{et al.}'s results using the entire collection of artist and genre, our results for collection style transfer.  }
	\label{fig:vs_gatys}
\end{figure*}

Figure \ref{fig:vs_gatys} reports the difference of methods. We can see that Gatys et al. \cite{gatys2016image} requires manually picking target style images that closely match the desired output. If the entire collection is used as target images, the transferred style is the average style of the collections. In contrast, our algorithm outputs diverse and reasonable images, each of which can be viewed as a sample from the distribution of the artist's style.

\subsubsection{Comparison with Universal Style Transfer}
\cite{li2017universal} aims to apply the style characteristics from a style image to content images in a style-agnostic manner. By whitening and coloring transformation, the feature covariance of content images could exhibit the same style statistical characteristics as the style images without requiring any style-specific training. 

\begin{figure}
	\centering
	\includegraphics[width=1\linewidth]{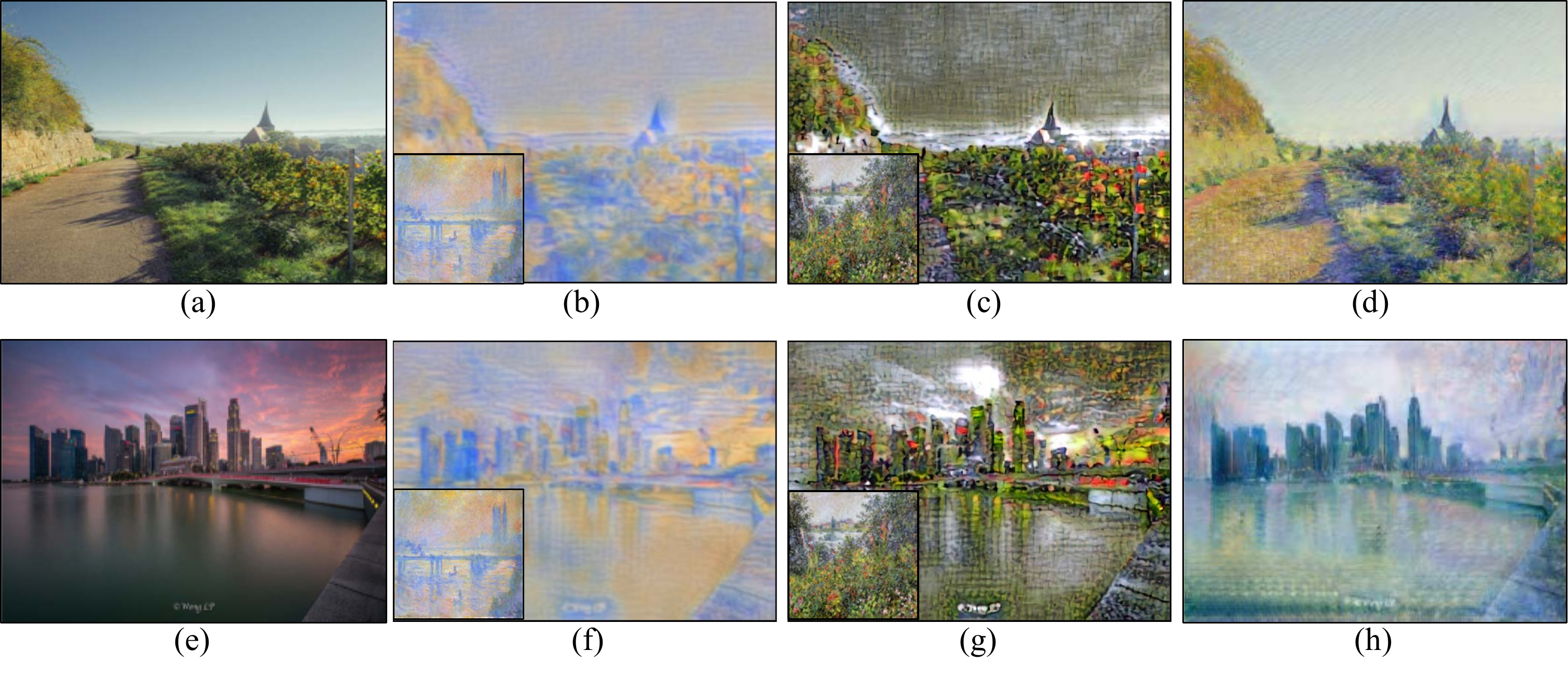}
	\vspace{-0.9cm}
	\caption{Comparison of our methods with universal style transfer \cite{li2017universal} on photo $\rightarrow$ Monet. From left to right: input images, results of \cite{li2017universal} with the style image: Monet \textit{Charing Cross Bridge}, results of \cite{li2017universal} with the style image: Monet \textit{Flowers at Vetheuil}, and our results of Monet's collection style transfer. }
	\label{fig:vs_uni}
\end{figure}

We compare images generated from the proposed algorithm and those from \cite{li2017universal}. The results are shown in Figure \ref{fig:vs_uni}. Given a picture with bushes and flowers (see Figure \ref{fig:vs_uni} (a)), our method outputs what Monet might record this scenery (see Figure \ref{fig:vs_uni} (d)), in which the style of painting bushes and flowers is similar to Monet’s painting of ``Flowers at Vetheuil''. What if the content image is a cityscape? Our method outputs images with foggy strokes (see Figure \ref{fig:vs_uni} (h)), since Monet produced a lot of cityscapes with fog in London (e.g. ``Charing Cross Bridge"). On the other hand, \cite{li2017universal} transfers images by following a particular style image. Taking ``Flowers at Vetheuil'' as the style image, Figure \ref{fig:vs_uni} (g) produced by \cite{li2017universal} well inherits the style of Monet’s ``Flowers at Vetheuil'' with green and red spot. However, Monet might not paint a cityscape with green and red spot as painting flowers. 

In summary, our task focuses on what the artists or genres might paint given content images, while the task of \cite{li2017universal} is to apply style characteristics from a particular style image to any content images. Both  \cite{li2017universal} and our method output interesting results, and could be used in different scenarios.

\subsubsection{Comparison with Collection Style Transfer}
\begin{figure*}
	\centering
	\includegraphics[width=0.9\linewidth]{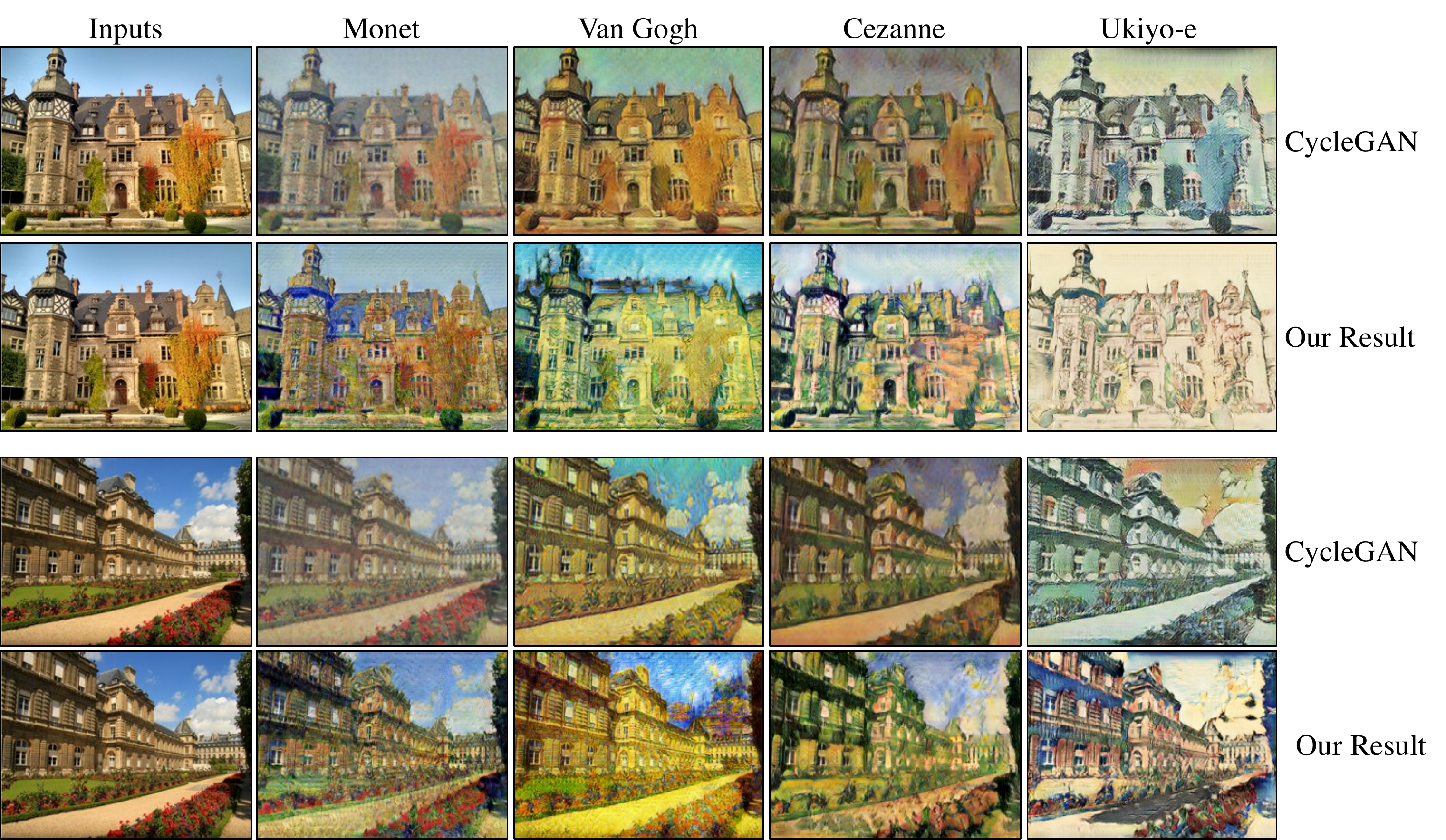}
	\caption{Comparison with CycleGAN \cite{zhu2017unpaired}. From left to right: original images, stylized images in Monet's style, stylized images in Van Gogh's style, stylized images in Cezanne's style, stylized images in Ukiyo-e style. In each case, the first row shows the results produced by CycleGAN, and
	the second row shows our results.}
	\label{fig:multivscyclegan}
\end{figure*}

CycleGAN \cite{zhu2017unpaired} previously showed impressive results on collection style transfer, so in this section we compare our results with CycleGAN. 
The generative network of baseline CycleGAN is composed of three stride-2 convolutional layers, 6 residual blocks, two fractional-convolutional layers and one last convolutional layer, which shares the same structure with our method in our experiment. 
Figure \ref{fig:multivscyclegan} demonstrates multi-collection style transfer by our method, which shows that the proposed model produces comparable results to CycleGAN. 

Quantitative results are shown in Table \ref{table:vs_cyclegan}, though the quality of images generated from the proposed algorithm exhibits similar performance as those of CycleGAN, it is instructive to note that our four styles are produced from a single network. In the second column of Table \ref{table:vs_cyclegan}, we compute the score of the corresponding content images of stylized images. We find that the stylized images achieve better performance than original content images. It demonstrates that the stylized images are more similar to the real authentic work of artists, which is consistent with our intuitive expectation. 

\begin{table}[]
	\centering
	\caption{Quantitative evaluation on collection style transfer in terms of FID to measure the performance. Lower score indicates better quality.}
	\label{table:vs_cyclegan}
	\begin{tabular}{cccc}
		\toprule
		Style& Content Images   & CycleGAN [14] &Ours  \\ \hline
		Monet & 86.50   & 64.14 &   55.13   \\  
		Cezanne   &  186.73&   106.96   & 107.27 \\ 
		Van Gogh & 173.01  & 107.03&109.59   \\
		Ukiyoe&  195.25& 103.36&115.96     \\
		MEAN & 160.37   &95.37&96.99     \\ 
		\bottomrule
	\end{tabular}
\end{table}

\begin{figure}
	\centering
	\includegraphics[width=0.9\linewidth]{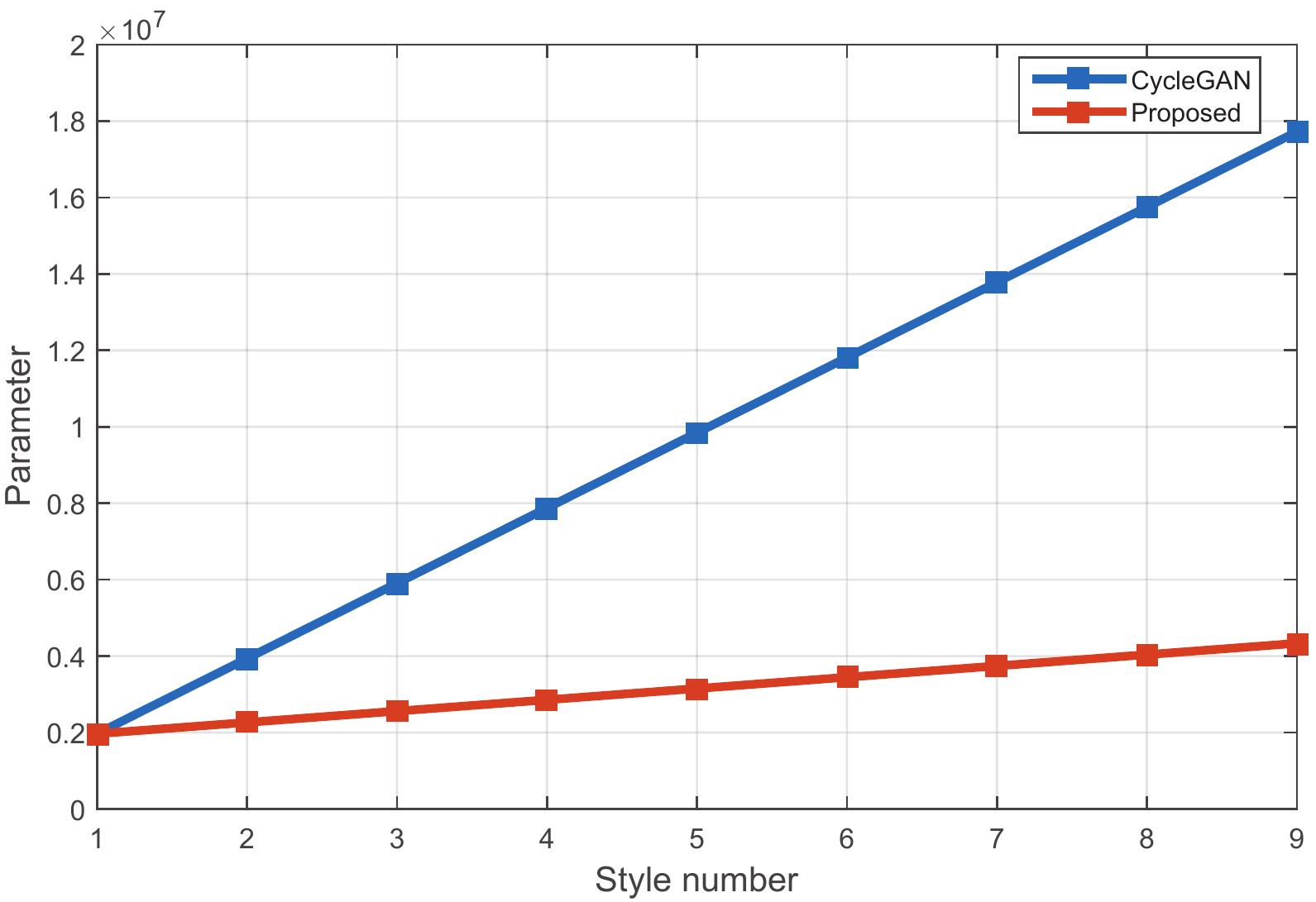}
	\vspace{-0.4cm}
	\caption{Model size.  We compare the number of parameters between our model and CycleGAN \cite{zhu2017unpaired}. The x-axis indicates style number and the y-axis indicates the model size.}
	\label{fig:vsmodelsize}
\end{figure}

Finally, we compare model size with CycleGAN \cite{zhu2017unpaired}. The generative network is composed of several convolutional layers and residual blocks with the same architecture as Gated-GAN when the transformer module number is set to one. The parameters of the two models are the same. Given another $N$ styles, CycleGAN must train another $N$ models. A whole generative network must be included for a new style. For Gated-GAN, the transformation operator is encoded in the gated transformer, which only has one residual block. A new style will thus only require a new transformer part in the generative network. As a result, the proposed method saves storage space as the style number increases. 
In Figure \ref{fig:vsmodelsize}, we compare the numbers of parameters with those of CycleGAN. Both  models are trained for $128 \times 128$ training images.

\subsubsection{Comparison with Conditional GAN}
Conditional GAN \cite{odena2016conditional,mirza2014conditional} model is a widely used method to generate classconditional image. When the conditional GAN is applied in multi style transfer, a stylized image $G(c,x)$ is generated from a content image $x$ and a style class label $c$. 
We compare conditional GAN in experiments. Class label is represented by a one-hot vector with $k$ bit where each represents a style type. $k$ noise vectors of the same dimension as the content image are randomly sampled from a uniform distribution. The input of generative network is obtained by concatenating the content image with the outer product of these noise vectors and the class label. 

\begin{figure}
	\centering
	\includegraphics[width=1\linewidth]{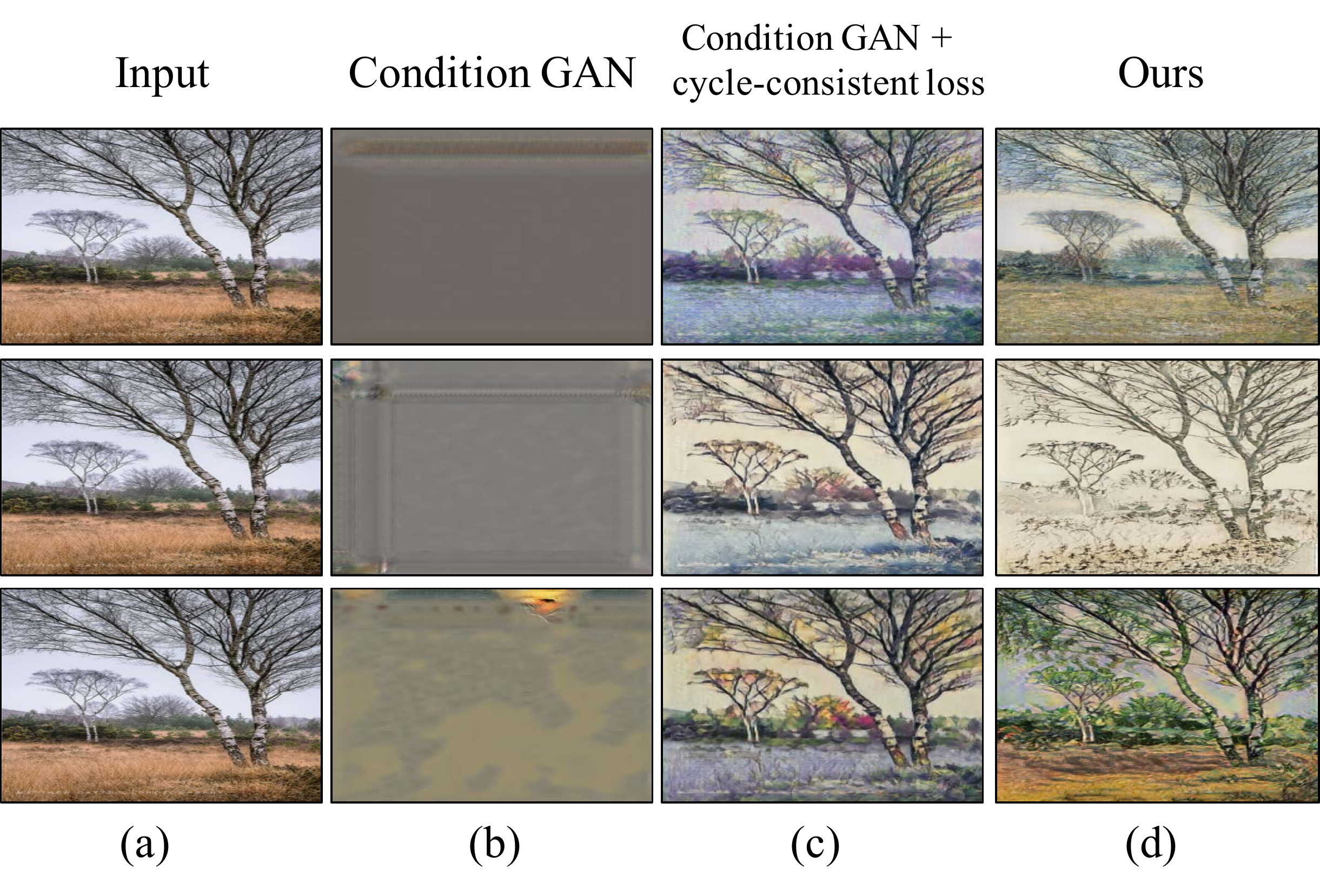}
	\vspace{-0.9cm}
	\caption{Comparison of our methods with Condition GAN and its variant. From left to right: input, condition GAN and condition GAN + cycle-consistent loss. Each row indicates different styles, from top to bottom: Monet, Ukiyo-e, Cezanne.}
	\label{fig:vs_condition}
\end{figure}

As we can see in Figure \ref{fig:vs_condition} (b), the conditional GAN fails to output meaningful results.This is because in collection style transfer, conditional GAN lacks of paired input-output examples. To stabilize the training of conditional GAN, we adopt cycle-consistent loss \cite{zhu2017unpaired}. From the results of conditional GAN with cycle-consistent loss in Figure \ref{fig:vs_condition} (c), we can see that the results of different styles tend to be much similar, and only colors are changed at first sight. In contrast, our results (see Figure \ref{fig:vs_condition} (d)) are more diverse in different styles in terms of strokes and textures. 
\subsection{Analysis of Loss Function}
\subsubsection{Influence of Parameters in Loss Function}
In our model, we proposed an auxiliary classier loss and an auto-encoder reconstruction loss, which are balanced by parameters $\lambda_{CLS}$ and $\lambda_R$ respectively. Now we analyze the influence of parameters. To explore the influence of parameters, we do experiments by considering $\lambda_{CLS}=\{0, 0.1, 1, 5, 10\}$ and $\lambda_R=\{1, 5, 10 ,20\}$. 

\begin{figure}
	\centering
	\includegraphics[width=0.98\linewidth]{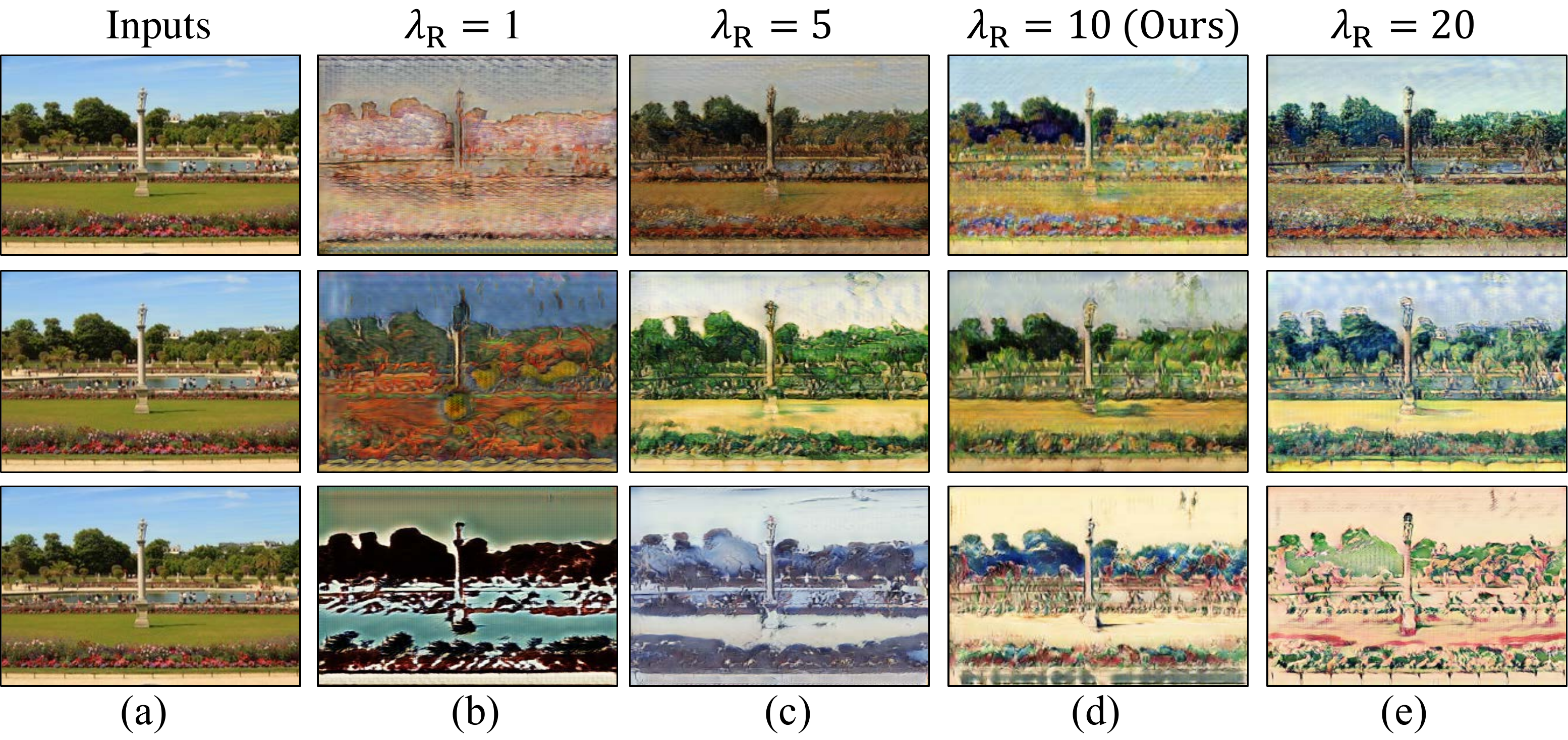}
	\vspace{-0.4cm}
	\caption{Qualitative comparison of the influence of parameter $\lambda_R$. The first column shows the input images. The rest columns demonstrate results with $\lambda_{R}=\{1,5,10,20\}$. Each row demonstrates images transferred by different styles. From top to bottom: Monet, Cezanne, Ukiyo-e.}
	\label{fig:lambda_r}
\end{figure}

Figure \ref{fig:lambda_r} and Table \ref{table:lambda_r} demonstrate the qualitative and quantitative comparisons of the influence of parameter . We can see the classifier loss provides a supervision of styles. Without classifier loss ($\lambda_{CLS}=0$), our model will only transfer into one style. If we set  too large ($\lambda_{CLS}=10$), the model would produce images with some artifacts. The underlying reason is that larger suppresses the function of discriminative network so that the output becomes less realistic. As a result, we set $\lambda_{CLS}=1$ in our model.
\begin{figure}
	\centering
	\includegraphics[width=0.98\linewidth]{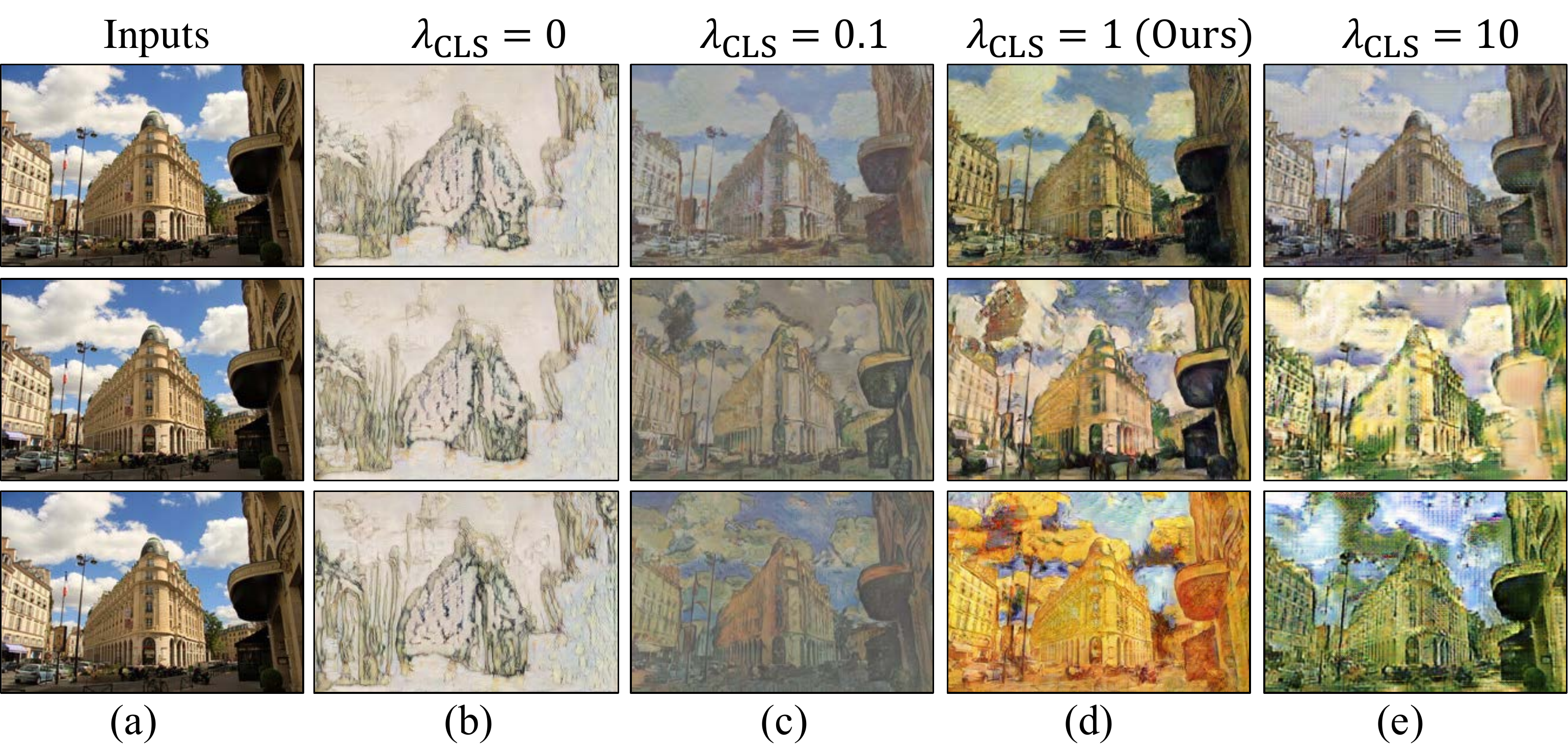}
	\vspace{-0.4cm}
	\caption{Qualitative comparison of the influence of parameter $\lambda_{CLS}$. The first column shows the input images. The rest columns demonstrate results with $\lambda_{CLS}=\{0,0.1,1,10\}$. Each row demonstrates images transferred by different styles. From top to bottom: Monet, Cezanne, Van Gogh.}
	\label{fig:lambda_cls}
\end{figure}

\begin{table}[]
	\centering
	\caption{Quantitative evaluation of parameter $\lambda_{CLS}=\{0,0.1,1,5,10\}$ in terms of FID score.}
	\label{table:lambda_cls}
	\scalebox{0.82}{
		\begin{tabular}{cccccc}
			\toprule
			Style& $\lambda_{CLS}=0$  & $\lambda_{CLS}=0.1$ & $\lambda_{CLS}=1$ & $\lambda_{CLS}=5$ & $\lambda_{CLS}=10$  \\ \hline
			Monet & 204.82&63.35&55.13&62.66&61.48\\  
			Cezanne   & 234.02&136.35&107.27&127.77&143.39 \\ 
			Van Gogh & 217.10&112.61&109.59&126.56&138.66\\
			Ukiyoe&  206.67&138.13&115.96&	132.72&	140.53\\
			MEAN & 215.65&112.61&96.99&112.42&121.02\\ 
			\bottomrule
	\end{tabular}}
\end{table}

Figure \ref{fig:lambda_r} and Table \ref{table:lambda_r} reveal the qualitative and quantitative comparisons of the influence of parameter $\lambda_R$. We can see that if we set $\lambda_R$ too small ($\lambda_R=1$), the outputs tend to be blurry and meaningless. It is because $\lambda_R$ improve the stability of training procedure. 
When $\lambda_R=\{5,10,20\}$ the visual qualities are similar while the FID score shows  achieves a slightly better quantitative performance. It demonstrates that our method is robust and easy to reproduce satisfying results. Since $\lambda_R=10$ achieves the best quality, we set $\lambda_R=10$ in our model.

\begin{table}[]
	\centering
	\caption{Quantitative evaluation of parameter $\lambda_{R}=\{1,5,10,20\}$ in terms of FID score.}
	\label{table:lambda_r}
	\begin{tabular}{ccccc}
		\toprule
		Style& $\lambda_{R}=1$  & $\lambda_{R}=5$ & $\lambda_{R}=10$ & $\lambda_{R}=20$ \\ \hline
		Monet & 180.30&121.07&55.13&115.09\\  
		Cezanne   & 165.27&148.67&107.27&140.84 \\ 
		Van Gogh & 148.43&139.87&109.59&134.13\\
		Ukiyoe&  166.69&134.26&115.96&	138.54\\
		MEAN & 165.17&135.97&96.99&132.15\\ 
		\bottomrule
	\end{tabular}
\end{table}
\subsubsection{Analysis of Auto-encoder Reconstruction Loss}
We next justify our choice of L1-norm. Beyond L1-norm, L2-norm can also be used in Equation \ref{autoencoder loss}. In Table \ref{table:L1}, we find that there is no significant difference between results of L1 and L2 loss. In CycleGAN [14], L1-norm is used in cycle-consistent reconstruction loss. As CycleGAN is an important comparison algorithm in our paper, we adopt L1-norm in our auto-encoder reconstruction loss as well.
\begin{table}[]
	\centering
	\caption{Quantitative evaluation on L1-norm and L2 norm in terms of FID score.}
	\label{table:L1}
	\begin{tabular}{cccc}
		\toprule
		Style&CycleGAN&	Ours (L1-norm)&	Ours (L2-norm)  \\ \hline
		Monet &64.14&	55.13&	56.09   \\  
		Cezanne   & 106.96&	107.27	&101.54 \\ 
		Van Gogh &107.03&	109.59	&109.33   \\
		Ukiyoe&103.36&	115.96&	112.39    \\
		MEAN & 95.37&	96.99&	94.84    \\ 
		\bottomrule
	\end{tabular}
\end{table}

\begin{figure*}[!h]
	\centering
	\includegraphics[width=0.95\linewidth]{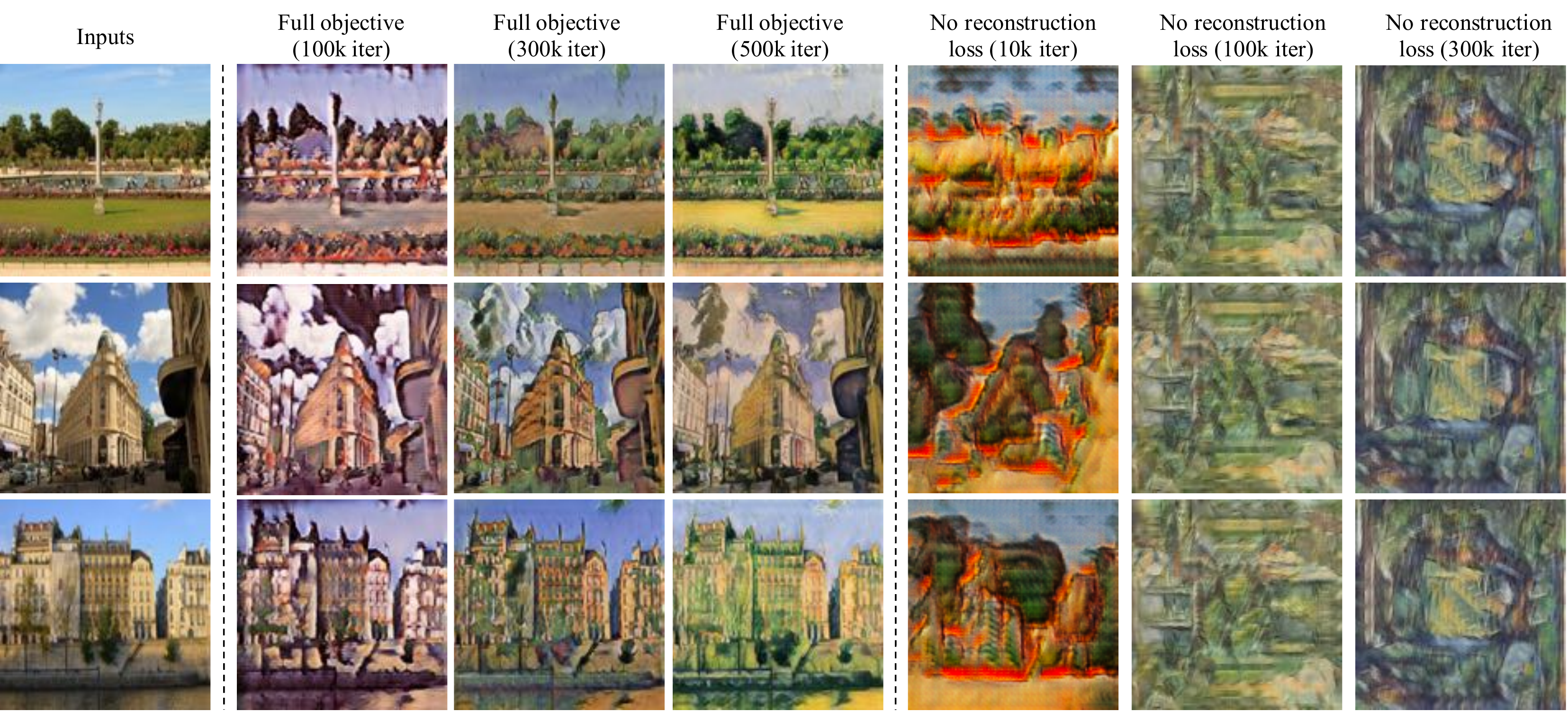}
	\caption{Comparison with a variant of our method across different training iterations for mapping images to Cezanne’s style. From left to right: original images, results after training for 10k, 100k, and 300k iterations with and without auto-encoder reconstruction loss.}
	\label{fig:ablation}
\end{figure*}

Lastly, we analyze the influence of the \textit{auto-encoder reconstruction loss} in stabilizing the adversarial training procedure.  
We train a comparative model by ignoring the \textit{auto-encoder reconstruction loss} in Equation \ref{autoencoder loss}. In Figure \ref{fig:ablation}, the model without Equation \ref{autoencoder loss} generates images with random texture and tend to be less diverse after training for several iterations. In contrast, the full proposed model generates satisfying results. 
Without the \textit{auto-encoder reconstruction loss}, the network only aims to generate images to fool the discriminative network, which often leads to the well-known problem of mode collapse \cite{arjovsky2017towards}. Our encoder-decoder subnetwork is encouraged to reconstruct input images, and thus semantic structure of the input is aligned with that of the output, which directly encourages diversity of output along with different inputs.
As a result, the full proposed model outputs satisfying results.

\subsection{Analysis of network architecture}
\begin{figure}
	\centering
	\includegraphics[width=1\linewidth]{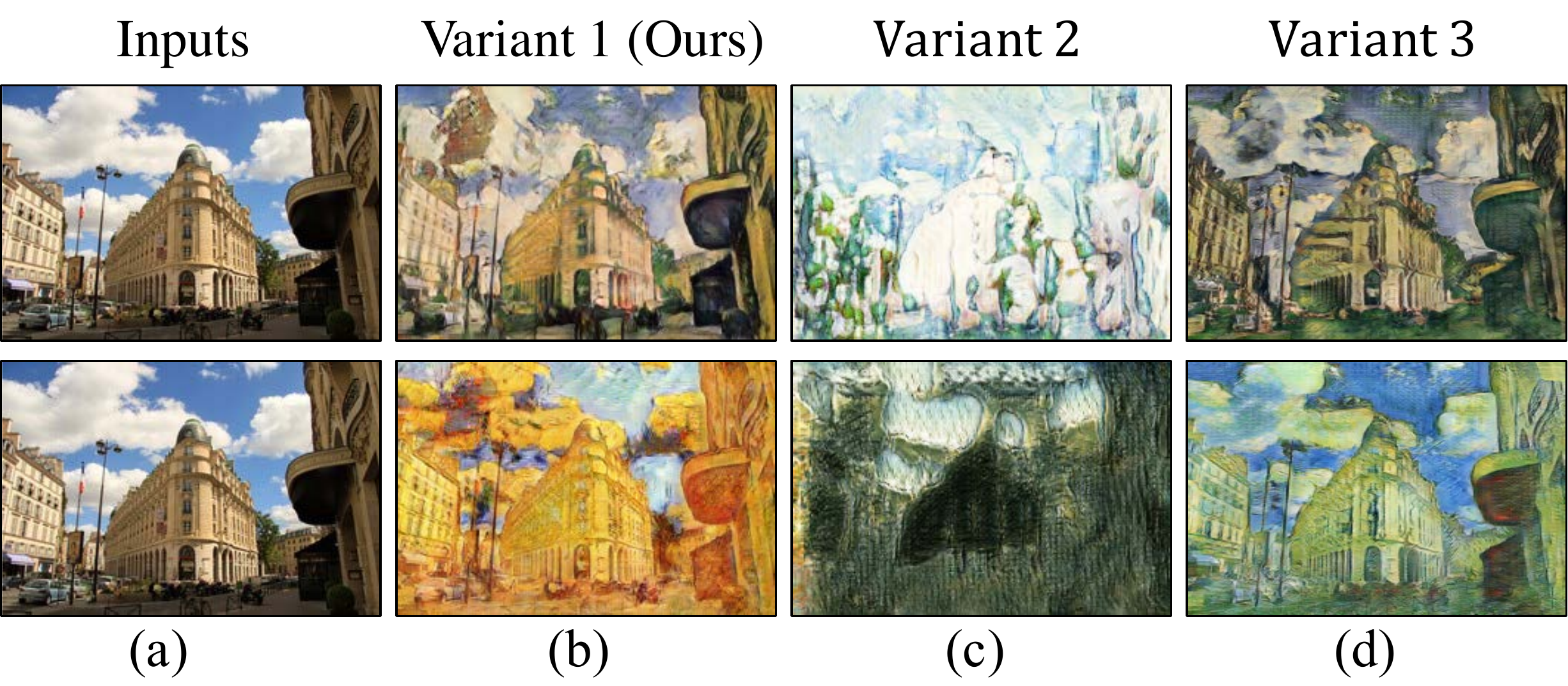}
	\caption{Qualitative comparison of the influence of different network structures. The first row is the results of photo $\rightarrow$ Cezanne, and the second row is the results of photo $\rightarrow$ Van Gogh.}
	\label{fig:vs_variant}
\end{figure}

We explore the influence of neural network structure. 
We setup variants of our model in Table \ref{table:arch}. Variants of models have different configurations of gated-transformer module. The quantitative results in Table \ref{table:restults_arch} reveal that the performance of variant 2 declines compared to that of the variant 1. From qualitative results in Figure \ref{fig:vs_variant}, we observe that model of variant 2 cannot maintains content structure (see Figure \ref{fig:vs_variant} (c)). The underlying reason is that the residual block has a branch that skip the convolutional layer and directly connects between the encoder and decoder module. Since the encoder-decoder subnetwork learns the content information of input from reconstruction loss, residual blocks with skipping connection shuttle the encoded information to the decoder module, which helps our model to output results aligned with the structured of input images.

To analyze the influence of layer size of gated-transformer module, we set variant 3 whose gated-transformer consists of 2 residual blocks. In Table \ref{table:restults_arch}, we can see the model of variant 3 achieves a slightly better quantitative evaluation than variant 1. The reason is that with the number of residual blocks increasing, the expression capacity of network increases as well, which means the model could capture more details for each style. However, the performance rise of variant 3 is limited and the qualitative qualities are similar in Figure \ref{fig:vs_variant}, which means one residual block of gated-transformer is sufficient in multi style transfer. As a result, we adopt variant 1 of architecture as our method.

\begin{table}
	\centering
	\caption{Experiment setup of network architecture analysis}
	\label{table:arch}
	\scalebox{0.67}{
		\begin{tabular}{|c|c|c|c|}
			\hlinew{0.8pt}
			& Expt1              & Expt2 & Expt3   \\ \hline
			Encoder & 3 $\times$ Convolution            & 3 $\times$ Convolution          &  3 $\times$ Convolution   \\ \hline
			Gated-transformer        & 1 $\times$ Residual block         & 1 $\times$ Convolution            &   2 $\times$ Residual block   \\ \hline
			\multirow{3}{*}{Decoder} & 5 $\times$ Residual block          &   5 $\times$  Residual block           &     5 $\times$ Residual block    \\
			& 2 $\times$Fractional-convolution &   2 $\times$ Fractional-convolution      & 2 $\times$ Fractional-convolution     \\
			& 1 $\times$ Convolution            & 1 $\times$ Convolution          &1 $\times$ Convolution  \\ 
			\hlinew{0.8pt}
	\end{tabular}}
\end{table}
\begin{table}
	\centering
	\caption{Quantitative evaluation on different network structure in terms of FID.}
	\label{table:restults_arch}
	\begin{tabular}{cccc}
		\toprule
		Style&Variants 1(Ours)&	Variants 2&	Variants 3  \\ \hline
		Monet &55.13&	67.16&	53.08   \\  
		Cezanne   & 107.27&	128.62&	110.13 \\ 
		Van Gogh &109.59&	199.71&	109.07   \\
		Ukiyoe& 115.96&	195.87&	100.32     \\
		MEAN &96.99&	147.84&	93.15    \\ 
		\bottomrule
	\end{tabular}
\end{table}

\subsection{Incremental Training}
By sharing the same encoding/decoding subnets, our model is compatible to the new style. For a new style, our model enables to add the style by learning a new branch in the gated-transformer while holding the encoding-decoding subnets fixed. 
We first jointly train the encoder-decoder subnetwork and gated-transformer (three collection style: Cezanne, Ukiyo-e and Van Gogh) with the strategy described in Algorithm \ref{alg:Framwork}. After that, for new the style (Monet), we train a new branch of residual blocks in the gated-transformer.

\begin{figure}
	\centering
	\includegraphics[width=1\linewidth]{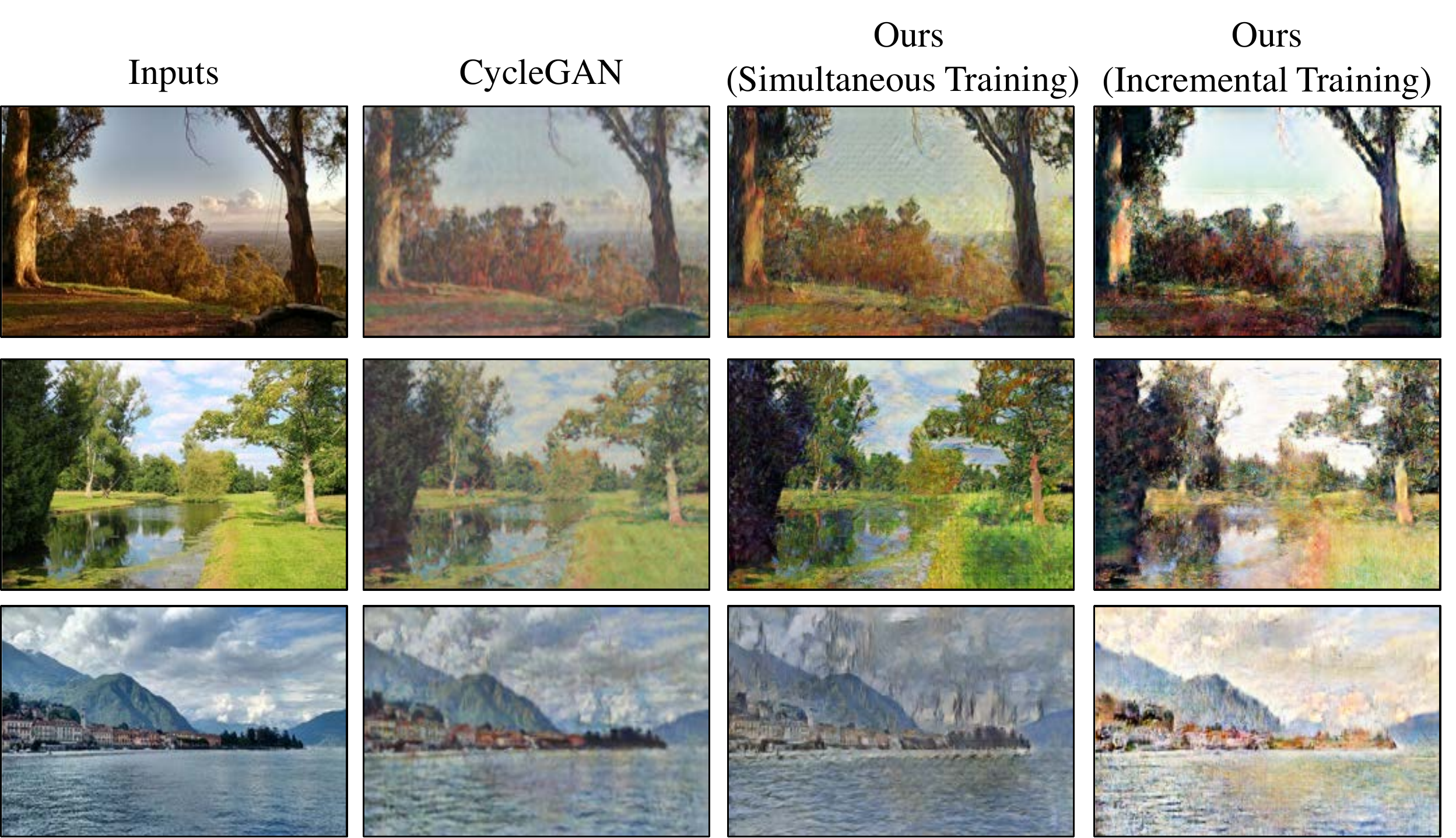}
	\caption{Comparison of incremental training. From left to right: original inputs, results of CycleGAN [14], results of our methods that all the styles are trained simultaneously, results of incremental training.}
	\label{fig:incre}
	\vspace{-0.3cm}
\end{figure}

Figure \ref{fig:incre} shows several results of new style by incremental training. It obtains very comparable stylized results to the CycleGAN, which trains the whole network with the style. We also evaluate the quantitative performance of the new style in term of FID score. The new style by incremental training gets score of 57.27. Compared to 55.13 of our Gated-GAN and 64.14 of baseline CycleGAN, the incremental training achieves a competitive result. 
 
\subsection{Linear Interpolation of Styles}
\begin{figure*}
	\centering
	\includegraphics[width=0.8\linewidth]{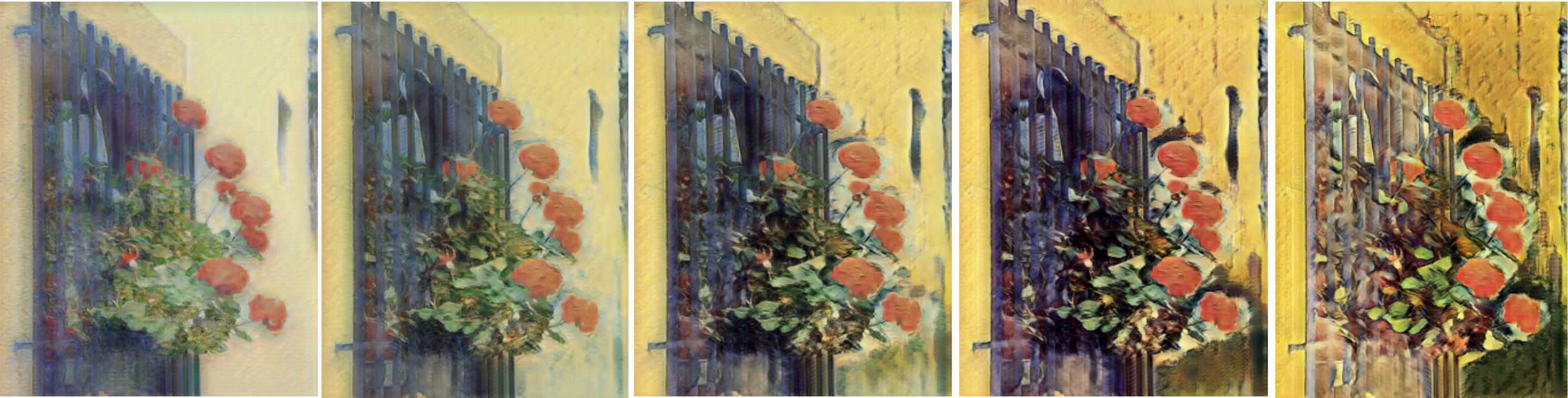}
	\caption{Style interpolation. The leftmost image is generated in Monet's style, and the rightmost image is generated in Van Gogh's style. Images in the middle are convex combinations of the two styles.}
	\label{fig:interpolation}
\end{figure*}

Since our proposed model achieves multi-collection style transfer by switching gates $c$ to different branches $T(Enc(x),c)$, 
we can blend multiple styles by adjusting the gate weights to create a new style or generate transitions between styles of different artists or genres:
\begin{equation}\label{eq:interpolation}
\tilde{G}(x,c_1,c_2)=Dec( \alpha \cdot T(Enc(x), c_1) + (1-\alpha)\cdot T(Enc(x), c_2))
\end{equation}
where $c_1$ and $c_2$ indicate the gates corresponding to different style branches, and  indicates the weight for convex combination of styles. 
In Figure \ref{fig:interpolation}, we show an example of interpolation from Monet to Van Gogh with the trained model as we vary $\alpha$ from $0$ to $1$.
The convex combination produces a smooth transition from one style to the other.

\section{Conclusions and Future Work}
 In this paper, we study multi-collection style transfer in a single network using adversarial training. To integrate styles into a single network, we design a gated network that filters in different network branches with respect to different styles. To learn multiple styles simultaneously, a discriminator and an auxiliary classifier distinguish authentic artworks and their styles. To stabilize GAN training, we introduce the auto-encoder reconstruction loss. Furthermore, the gated transformer module provides the opportunity to explore new styles by assigning different weights to the gates. Experiments demonstrate the stability, functionality, and effectiveness of our model and produce satisfactory results compared with a state-of-art algorithm, in which one network merely outputs images in one style.
In the future, we will apply our model to train other conditional image generation tasks (e.g., object transfiguration, season transfer, photo enhancement) and explore to generate diversified style transfer results.

\appendices

\ifCLASSOPTIONcaptionsoff
  \newpage
\fi

\bibliographystyle{IEEEtran}
\bibliography{bibs}

\end{document}